\documentclass[10pt]{article} % For LaTeX2e
\usepackage[accepted]{tmlr}
\usepackage{amsmath,amsfonts,bm}

\def\eqref#1{equation~\ref{#1}}
\def\1{\bm{1}}

\DeclareMathAlphabet{\mathsfit}{\encodingdefault}{\sfdefault}{m}{sl}
\SetMathAlphabet{\mathsfit}{bold}{\encodingdefault}{\sfdefault}{bx}{n}

\usepackage{hyperref}
\usepackage{url}

\usepackage{hyperref}       % hyperlinks
\usepackage{url}            % simple URL typesetting
\usepackage{booktabs}       % professional-quality tables
\usepackage{amsfonts}       % blackboard math symbols
\usepackage{nicefrac}       % compact symbols for 1/2, etc.
\usepackage{microtype}      % microtypography
\usepackage{xcolor}         % colors

\usepackage{wrapfig}
\newcommand{\model}{\texttt{PI-GNN}\xspace}
\usepackage{graphicx}
\usepackage{amsmath}
\usepackage{amssymb}
\usepackage{enumitem}
\usepackage{subcaption}	% You need this package for subfigures
\usepackage{booktabs}
\usepackage{multirow}
\usepackage{xspace}
\usepackage{color, colortbl}
\definecolor{Gray}{gray}{0.9}

\def\*#1{\mathbf{#1}}
\usepackage{algorithm}
\usepackage[ruled,vlined,algo2e]{algorithm2e}

\title{Noise-robust Graph Learning by Estimating and Leveraging Pairwise Interactions}

\author{\name Xuefeng Du\thanks{Xuefeng and Tian contributed equally to this work. Work is done while interning at Tencent.} \email xfdu@cs.wisc.edu \\
      \addr 
      University of Wisconsin-Madison
      \AND
      \name Tian Bian$^*$ \email tbian@se.cuhk.edu.hk \\
      \addr 
The Chinese University of Hong Kong
      \AND
      \name Yu Rong \email  yu.rong@hotmail.com\\
      \addr Tencent AI Lab 
\AND
      \name Bo Han \email  bhanml@comp.hkbu.edu.hk\\
      \addr Hong Kong Baptist University
         \AND
      \name
         Tongliang Liu \email tongliang.liu@sydney.edu.au \\
         \addr Mohamed bin Zayed University of Artificial Intelligence \\
         The University of Sydney 
         \AND
      \name Tingyang Xu \email  yu.rong@hotmail.com\\
      \addr Tencent AI Lab 
      \AND
      \name Wenbing Huang \email hwenbing@126.com \\
      \addr Renmin University of China 
        \AND
      \name Yixuan Li \email sharonli@cs.wisc.edu \\
      \addr University of Wisconsin-Madison 
  \AND
      \name
      Junzhou Huang \email  jzhuang@uta.edu  \\ 
      \addr  University of Texas at Arlington 
}

\def\openreview{\url{https://openreview.net/forum?id=r7imkFEAQb}} % Insert correct link to OpenReview for camera-ready version

\begin{document}
 % we observe that Pairwise Interaction (PI), , is a useful learning proxy in the context of noisy node labels. 

\maketitle

\begin{abstract}
Teaching Graph Neural Networks (GNNs) to accurately classify nodes under severely noisy labels is an important problem in real-world graph learning applications, but is currently underexplored. Although \emph{pairwise} training methods have
demonstrated promise in supervised metric learning and unsupervised contrastive learning, they remain less studied on noisy graphs, where the structural pairwise interactions (PI) between nodes are abundant and thus might benefit label noise learning rather than the \emph{pointwise} methods. This paper bridges the gap by proposing a pairwise framework for noisy node classification on graphs, which relies on the PI as a primary learning proxy in addition to the pointwise learning from the noisy node class labels. Our proposed framework \model contributes two novel components: (1) a confidence-aware PI estimation model that adaptively estimates the PI labels, which are defined as whether the two nodes share the same node labels, and (2) a decoupled training approach that leverages the estimated PI labels to regularize a node classification model for robust node classification. Extensive experiments on different datasets and GNN architectures demonstrate the effectiveness of \model, yielding
a promising improvement over the state-of-the-art methods.  Code is publicly available at \url{https://github.com/TianBian95/pi-gnn}.

% Semi-supervised node classification, as a fundamental problem in graph
% learning, leverages unlabeled nodes along with a small portion of labeled nodes for training. Existing methods rely heavily on high-quality labels, which are expensive to obtain in real-world applications since certain noises are inevitably involved during the labeling process, especially when the dataset scales up. It hence poses an unavoidable challenge for the learning algorithm to generalize well. 

% In this paper, we propose a novel robust learning paradigm that estimates and leverages the Pairwise Interactions (PI) for the model, such as  to combat noisy node labels. Unlike \emph{pointwise} robust training approaches, PI-GNN explicitly shapes the embeddings of node pairs by the \emph{pairwise interaction} (PI) label, which can be applied to both labeled and unlabeled nodes. We design several instantiations for PI labels based on the graph structure and the node class labels, and further propose a new confidence-aware contrastive training technique to mitigate the negative effect of the sub-optimal PI labels. Extensive experiments on different datasets and GNN architectures demonstrate the effectiveness of PI-GNN, yielding
% a promising improvement over the state-of-the-art methods.

\end{abstract}

\section{Introduction}
Graphs are ubiquitously used to represent data in different fields, including social networks, bioinformatics, recommendation systems, and computer network security. Accordingly, graph analysis tasks, such as node classification, have a significant impact in reality~\citep{DBLP:conf/iclr/XuHLJ19}. The success of machine learning models, such as graph neural networks (GNNs) on node classification relies heavily on the collection
of large datasets with human-annotated labels~\citep{DBLP:conf/cikm/0002CZTZG19}.  However, it is extremely expensive and time-consuming
to label millions of nodes with high-quality annotations. Therefore, when dealing with large graphs, usually a subset of nodes is labeled, and a wide spectrum of semi-supervised learning techniques have emerged for improving node classification performance~\citep{DBLP:conf/icml/ZhuGL03,DBLP:conf/nips/ZhouBLWS03,DBLP:conf/iclr/KipfW17}.

Although achieving promising results, these techniques overlook the existence of noisy node labels. For instance, practitioners often leverage inexpensive alternatives for annotation, such as combining human and machine-generated label~\citep{DBLP:conf/nips/HuFZDRLCL20}, which inevitably yields samples with noisy labels. Since neural networks (including GNNs) are able to memorize any given (random) labels~\citep{DBLP:journals/corr/abs-1905-01591,DBLP:conf/iclr/ZhangBHRV17}, these noisy labels would easily prevent them from generalizing well. Therefore, training robust GNNs for \emph{semi-supervised node classification against noisy labels} becomes increasingly \emph{crucial but less studied} for safety-critical graph analysis, such as predicting the identity groups of users in social networks or the function of proteins to facilitate wet laboratory experiments, etc.

\begin{figure*}[!htp]
    \centering
    % \vspace{-5mm}
    \includegraphics[width=0.8\textwidth]{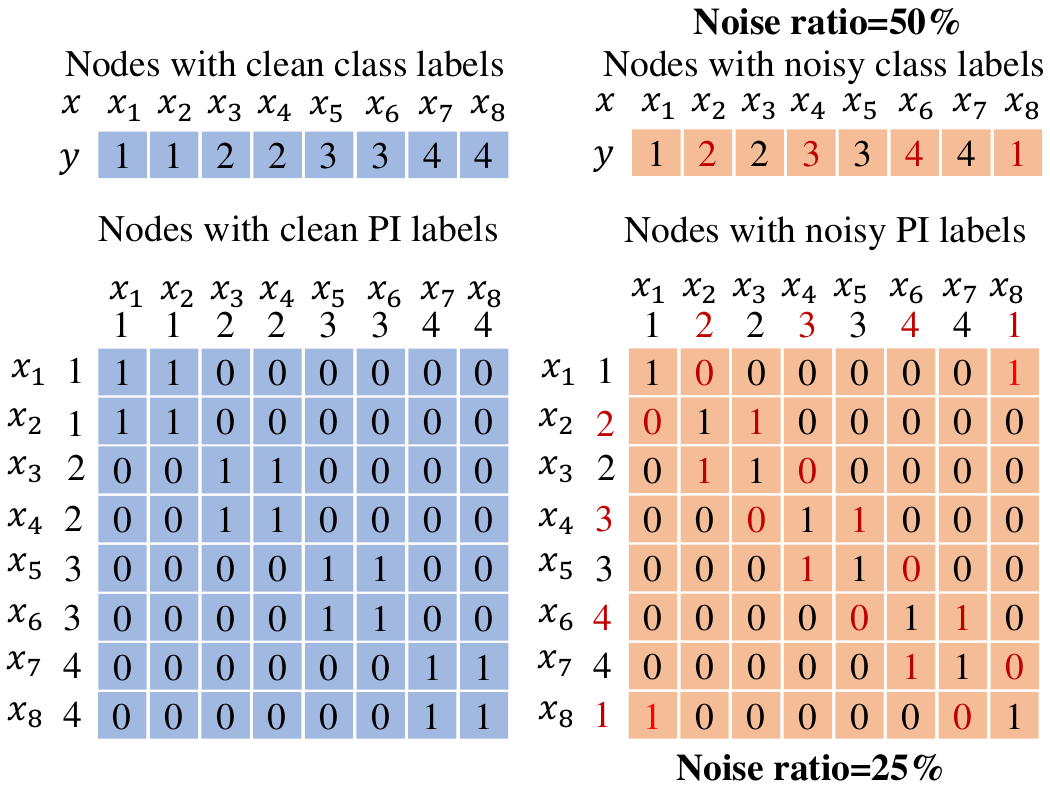}
     % \vspace{-1.1em}
     \caption{\small \textbf{Noise ratio comparison with noisy node labels}. The noise ratio of PI labels is much smaller than that of node labels. Number in red denotes noisy label. 
     }
       \label{fig:motivation}
    % \vspace{-1.9em}
\end{figure*}
%Analytical results on CiteSeer by a two-layer GNN.
% which justifies the \emph{pairwise interactions} and \emph{uncertainty estimation} in PI-GNN.

%The bar plot shows the average result from 10 different trials.

%(b) A sample density plot for the confidence mask, which shows there exist unconfident PI predictions. x-axis denotes the confidence value. (c) Test accuracy of the GNN trained by the PI loss from all node pairs (Uniform), the node pairs with larger (Larger Confi.) and smaller confidence (Smaller Confi.), all node pairs weighted by the confidence mask (CPL). Models trained on node pairs with higher confidence perform much better than those trained with lower confidence or without PI labels. PI label is instantiated as the adjacency matrix here.
In this paper, we pioneer a pairwise framework for noise-robust node classification on graphs, where relationships
between data points are exploited. Currently, although the pairwise approaches are prevailing and made great progress in supervised metric learning and unsupervised
contrastive learning~\citep{qi2020simple,boudiaf2020unifying,chen2020simple,he2020momentum}, they remain largely unexplored in noise-robust graph learning. In particular, existing pointwise noise-robust learning algorithms~\citep{DBLP:journals/corr/abs-1905-01591,xia2021towards,DBLP:journals/corr/abs-2103-03414,DBLP:journals/corr/abs-2010-12408,DBLP:journals/corr/abs-2012-12896} are mainly designed for image inputs and strictly rely on the class label that shows the class that a node belongs
to for learning. In contrast, the pairwise framework is able to utilize the pairwise interactions (PI) between nodes, which indicate whether or not two
nodes belong to the same class, as a learning proxy. As a result, it reduces the multi-class classification problem into a binary classification
problem, which is easier
to handle~\cite{DBLP:conf/cvpr/PatriniRMNQ17} and provides helpful learning signals apart from  the noisy pointwise supervision. \textcolor{black}{For example, Figure~\ref{fig:motivation}\footnote{\textcolor{black}{There are some additional cases where the conclusion might not hold true, please see Appendix Section~\ref{sec:app_corner_case}}}. shows the transformation from the class labels to the PI labels.} We can easily observe that the noise rate for the PI labels is much lower than that of the pointwise noisy class labels.  Considering two nodes from the same class have the same noisy labels, their PI label still remains positive, which is helpful for the model to learn.

Although learning with PI intuitively demonstrates promise, it does not
trivially transfer to label noise learning on graphs. For example, previous pairwise learning frameworks~\citep{qi2020simple,chen2020simple} can easily calculate the PI labels either through class label comparison (same class label$\rightarrow$positive PI label) or data augmentation (augmented views from the same image$\rightarrow$positive PI label). However, PI labels can still contain unneglectable noise (\emph{cf.} Figure~\ref{fig:motivation}) if we directly compare their noisy node class labels. As a result, the pairwise learning algorithm relying on
such suboptimal PI labels can misbehave.

We propose a novel framework dubbed \model, tackling two highly dependent problems—PI estimation and learning—in one synergistic framework. Concretely, \model contributes two novel
components: \textbf{(1)} We introduce an end-to-end confidence-aware PI label estimation branch that dynamically estimates PI labels with the help of graph structure (Section~\ref{sec:pi_gnn}). In particular, we learn a graph neural network that is trained to predict node connectivity, where the connected nodes have a ground truth of 1 and vice versa. Compared to using node connectivity as the PI label directly, \emph{i.e.}, connected nodes transform to a positive PI label, we derive PI labels with the predictive confidence from a PI label estimation network to quantify the reliability of such graph structure. \textbf{(2)} We explore a novel decoupled training approach by leveraging the estimated PI labels for learning a node classification model to perform noise-robust node classification (Section~\ref{sec:learning}). We propose to decouple the PI label estimation procedure from training with noisy node labels to prevent corruption on the estimated PI labels. Meanwhile, different from previous works~\cite{li2021unified}, our \model does not require a clean set of node and label pairs as extra supervision and can  simultaneously utilize both the labeled and unlabeled nodes for training, which works well for semi-supervised node classification.

Our main contributions are summarized as follows: 
\begin{itemize}
    \item We propose to train robust GNNs against noisy labels for node classification, which serve as a crucial step towards the reliable deployment of GNNs in complex real-world applications.
    \item We introduce a novel learning framework to simultaneously estimate and leverage the pairwise interactions, which can be applied on both labeled and unlabeled nodes without  extra supervision of clean node labels. 
    \item We demonstrate \model can be effectively used on different datasets, GNN architectures and different noise types and rates, \emph{e.g.}, improving the test accuracy by 5.4\% on CiteSeer under a severe label noise.
\end{itemize}

\section{Preliminaries}
\textbf{Graph Neural Networks.} Let $G=(V,E, A)$ be a graph with node feature vectors $X_v$ for $v\in V$ and edge set $E$, where $A$ denotes the adjacency matrix. GNNs use the graph structure and node features $X_v$ to learn a representation vector of a node $h_v$, or the entire graph $h_{G}$, which usually follow a neighborhood aggregation strategy and iteratively update the representation of a node by aggregating representations
of its neighbors. After $k$ iterations of aggregation, a node’s representation captures the structural
information within its $k$-hop network neighborhood. Formally, the $k$-th layer of a GNN is
% \vspace{-0.5em}
\begin{equation}
\small
\begin{aligned}
       a_{v}^{(k)}=\text{AGGREGATE }^{(k)}\left(\left\{h_{u}^{(k-1)}: u \in \mathcal{N}(v)\right\}\right),  h_{v}^{(k)}=\text{COMBINE}^{(k)}\left(h_{v}^{(k-1)}, a_{v}^{(k)}\right),
\end{aligned}
% \vspace{-1em}÷
\end{equation}
where $h_{v}^{(k)}$ is the feature vector of node $v$ at the $k$-th layer. $h_{v}^{(0)}=X_v$. $\mathcal{N}(v)$ denotes the neighboring nodes of $v$. The choices of $\text{AGGREGATE }^{(k)}(\cdot)$ and $\text{COMBINE}^{(k)}(\cdot)$ can be diverse among different GNNs. For example, in GCN~\cite{DBLP:conf/iclr/KipfW17}, the element-wise mean pooling is used, and the AGGREGATE and COMBINE steps are integrated as follows:
\begin{equation}
\small
    h_{v}^{(k)}=\operatorname{ReLU}\left(W \cdot \operatorname{MEAN}\left\{h_{u}^{(k-1)}, \forall u \in \mathcal{N}(v) \cup\{v\}\right\}\right),
    % \vspace{-0.5em}
\end{equation}
where $W$ is a learnable matrix. For node classification, each node $v\in V$ has an associated label $ y_v$, the node representation $h_{v}^{(K)}$ of the final layer is used for prediction.

\textbf{Label-noise representation learning for GNNs.} Let $X_v$ be the feature and $y_v$ be the label for node $v$, we deal with a dataset $\mathcal{D}=\{\overline{\mathcal{D}}^{\mathrm{tr}}, \mathcal{D}^{\mathrm{te}}\}$ which consists of training set $\overline{\mathcal{D}}^{\mathrm{tr}}=\{(A,X_v, \overline{y}_v)\}_{v\in V}$ that is drawn from a corrupted distribution $\overline{D}=p(A,X,\overline{Y})$ where $\overline{Y}$ denotes the corrupted label. Let $p(A,X,Y)$ be the non-corrupted joint probability distribution of features $X$ and labels $y$, and $f^{*}$ be the (Bayes) optimal hypothesis from $X$ to $y$. To approximate $f^{*}$, the objective requires a hypothesis space $\mathcal{H}$ of hypotheses $f_{\theta}(\cdot)$ parametrized by $\theta$. A robust algorithm against noisy labels contains the optimization policy to search through $\mathcal{H}$ in order to find $\theta^{*}$ that corresponds to the optimal function in the hypothesis for $\overline{\mathcal{D}}^{\mathrm{tr}}: f_{\theta^{*}}\in \mathcal{H}$, and meanwhile is able to assign correct labels for $\mathcal{D}^{\mathrm{te}}$.
\vspace{-0.7em}
\section{Proposed Approach}

\begin{figure*}[t]
    \centering
    \includegraphics[width=1.0\textwidth]{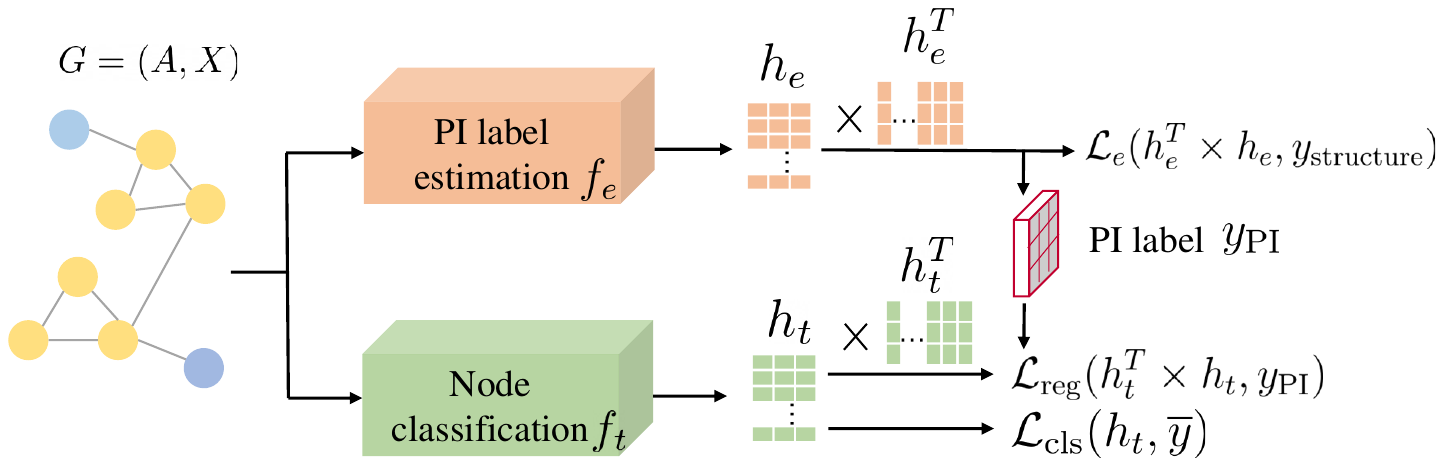}
    \caption{\small \textbf{The framework of our proposed \model}, which consists of two different branches, \emph{i.e.}, a PI label estimation branch and a node classification branch for noise-robust semi-supervised node classification. The two branches $f_e,f_t$ first estimate  the pairwise interactions between each node pair by the graph structure, and then leverage the estimated PI labels for joint training with the node classification task. $\times$ denotes the dot product operation. }%while $\scriptsize{f_{W_i^j}(\cdot)}$ means the atomic block with parameter $W_i^j$
    \label{fig:overview}
% \vspace{-1em}
\end{figure*}

%The mask generator generates a confidence mask, which is applied to the PI loss of the task executor to reduce the uncertainty of its predictions caused by the sub-optimal PI labels
In this section, we introduce our proposed \model, which performs noise-robust semi-supervised node classification by explicitly estimating and leveraging the pairwise interactions on graphs. In what follows, we will first provide a method overview and then illustrate the confidence-aware estimation of the pairwise interactions in \model (Section~\ref{sec:pi_gnn}). We introduce the decoupled training strategy for leveraging the estimated pairwise interactions for model  regularization (Section~\ref{sec:learning}).

\textbf{Overview.} Figure~\ref{fig:overview} demonstrates the overview of \model, which
%The overview of PI-GNN is shown in Figure~\ref{fig:overview}, which
is composed of two different branches. The confidence-aware PI label estimation branch takes in the graph structure and generates the estimated PI labels for a node pair. We denote it as $f_e$. 
% is only trained with the PI learning objective, whose outputs are used to generate the confidence mask. 
The node classification branch  takes the estimated PI labels and trains a node classification model jointly with an additional regularization objective, which leverages the PI labels to regularize the node embeddings. We denote it as  $f_t$. 

% is trained with both the PI learning objective and the noisy node classification objective, which is used to perform the node classification task. The PI loss for each node pair in the task executor is multiplied by the confidence mask from the mask generator in order to reduce the uncertainty caused by the sub-optimal PI labels.

\subsection{Confidence-aware PI Label Estimation}
\label{sec:pi_gnn}

Let us suppose certain node class labels in the training set $\overline{\mathcal{D}}^{\mathrm{tr}}=\{(A,X_v, \overline{y}_v)\}_{v\in V}$ are corrupted with noisy labels. Since, ultimately, we are interested in finding a GNN model $f$ parametrized by $\theta$ that minimizes the generalization error on a clean test set $\mathcal{D}^{\mathrm{te}}$, a natural solution is to exploit additional information during training for the learning algorithm to find a robust parameter $\theta^{*}$ in the hypothesis space $\mathcal{H}$. One straightforward candidate for such information is leveraging the pairwise interactions between two nodes to perform extra regularization, whose learning objective is shown to hold a much smaller noise rate than that with the noisy class labels (Figure~\ref{fig:motivation}).

\textbf{Train the PI label estimation model.} In this paper, we propose to estimate the PI labels $y_{\text{PI}}\in \mathbb{R}^{|V|\times |V|}$ between node pairs by learning from the graph structure (Here $|V|$ is the cardinality of the vertex set on the input graph $G$). While a reasonable choice of $y_{\text{PI}}$ is by comparing whether two nodes have the same class label $y$ and assigning those with the same class label a positive PI label, it is \emph{impossible} to obtain such PI labels with noisy class labels $\overline{y}$. 

Therefore, we propose to learn from the graph structure $A$ for estimating the PI labels using a PI label estimation model $f_e$ with paprameter $\theta_e$, such as a graph neural network. Specifically,   given node embeddings $h$ which is calculated by $h=f_e(A,X,\theta_e)$, assume $i,j\in V$, let $h_i^T\cdot h_j$ be the dot product of two node embeddings, the training objective for the PI label estimation model $\mathcal{L}_{e}\in \mathbb{R}^{|V|\times|V|}$ is formulated as follows:

% \vspace{-1em}
% \begin{equation}
% \small
%         \mathcal{L}_e\left(h ;G\right) =\sum_{(i,j) \in B^{+}}-\log P(h_i^T\cdot h_j)+ \lambda \cdot \sum_{(i,j) \in B^{-}}-\log (1-P(h_i^T\cdot h_j)),
%     \label{eq:pi_loss}
%     % \vspace{-0.4em}
% \end{equation}

\begin{equation}
% \color{blue}
    \mathcal{L}_e\left(h ;A,X\right)=\lambda \cdot \mathbb{E}_{{(i,j) \in B^{-}}}\left[-\log \frac{1}{1+\exp ^{h_i^T\cdot h_j}}\right]
    +\mathbb{E}_{{(i,j) \in B^{+}}}\left[-\log \frac{\exp ^{h_i^T\cdot h_j}}{1+\exp ^{h_i^T\cdot h_j}}\right],
\label{eq:pi_loss}
\end{equation}

%$y_{\text{structure}}$ is 1 if the node pair is connected and 0 if not. 
where $B^{+},B^{-}$ denote the node pairs that are connected and disconnected, respectively. $\lambda$ is applied for the disconnected node pairs to deal with the sample imbalance problem, which can be calculated according to~\cite{kipf2016variational,DBLP:conf/iclr/KipfW17}. 

\textbf{Estimate the PI labels.}  Given a well-trained PI label estimation model $f_e$, we derive the PI labels by taking the predictive confidence as the smoothed PI label, which is calculated as follows:
\begin{equation}
    y_{\text{PI}}(i,j) =\frac{\exp ^{h_i^T\cdot h_j}}{1+\exp ^{h_i^T\cdot h_j}}.
    \label{eq:estimate}
\end{equation}
\begin{algorithm}[H]
\SetAlgoLined
\textbf{Input:} Input graph $G=(V,A,X)$ with noisy training data $\overline{\mathcal{D}}^{\mathrm{tr}}=\left\{\left(A, X_{v}, \bar{y}_{v}\right)\right\}_{v \in V}$, randomly initialized GNNs $f_e$ and $f_t$ with parameter $\theta_e$ and $\theta_t$, weight for regularization loss $\beta$, pretraining epoch $K$ for $f_e$. Total training epoch $N$. \\
\textbf{Output:} Robust GNN $f_t$.\\
\For{$epoch=0; epoch < N; epoch++$}{
\eIf{epoch $\leq$ K}{
  Update the parameter $\theta_e$ of the PI label estimation model $f_e$ by Equation~\ref{eq:pi_loss}.\;
  
   Set $\beta=0$ in Equation~\ref{eq:all}, update the parameter $\theta_t$ of the node classification model $f_t$.
   }{
   Update the parameter $\theta_e$ of the PI label estimation model $f_e$ by Equation~\ref{eq:pi_loss}.\;
   
  Estimate the PI label $y_{\text{PI}}$ by Equation~\ref{eq:estimate} with $f_e$.\;
  
  % Calculate the PI loss $\ell_{PI}^t$ for the task executor $f_t$ by Equation~\eqref{eq:uncertainty}.
  
  Update the parameter $\theta_t$ of the node classification model $f_t$ by Equation~\ref{eq:all}.
  }
 }
 \Return The node classification model $f_t$.
 \caption{\model: Noise-robust Graph Learning by Estimating and Leveraging
Pairwise Interactions} 
 \label{alg:algo}
\end{algorithm}
where $y_{\text{PI}}(i,j)$  is proportional to the value of the dot product $h_i^T\cdot h_j$. The estimated PI label measures the predictive confidence by looking at the closeness between the prediction and the graph structure (\emph{i.e.,} node connectivity). If the predictive confidence becomes far away from the binary node connectivity (0 or 1), then the reliability of such predictions is relatively low, which results in a smoothed PI label and vice versa. We show in Section~\ref{sec:ablation} the ablations of training with different kinds of  PI labels, including training with the PI labels by comparing the clean node labels.
% and more attention should be paid for such node pairs. 

The PI label estimation procedure allows for explicitly exploiting the pairwise interactions between two nodes, resulting in a GNN that is affected less by the noisy class labels.~\cite{DBLP:journals/corr/abs-2006-07831} employed similarity labels for regularization. However, it transforms the noise transition matrix estimated from the noisy class labels $\overline{y}$ to correct the similarity labels, which is sensitive to the matrix estimation quality. Meanwhile, their approach did not deal with the noise-robust graph learning problem.

\subsection{Decoupled Noise-Robust Training}
\label{sec:learning}
Given the well-estimated PI labels, now we discuss how to leverage the PI labels for noise-robust training. One simple solution is to  train a single GNN to estimate the PI labels, leverage the estimated PI labels by replacing the binary labels $y_{\text{structure}}$ with the estimated smoothed PI labels and then perform regularization by  joint training with the node classification task on that single GNN. However, since the GNN is exposed to the noisy node class labels $\overline{y}$, \emph{the PI label $y_{\text{PI}}$ cannot be estimated well if we entangle both the node label prediction and PI label estimation in a single GNN}. 

% \textbf{Confidence estimation.} In order to train a robust GNN that is not sensitive to the sub-optimal PI labels, we resort to \textit{confidence estimation} and reduce the negative effect of the node pairs that the model is unconfident about during regularization (Equation~\eqref{eq:pi_loss}). Specifically, we measure the confidence of the PI predictions as follows:
% \begin{equation}
%     M(i,j) = \left\{\begin{array}{lr}
%         \sigma(h_i^{T}\cdot h_j), & y_{PI}(i,j)=1\\
%        1-\sigma(h_i^{T}\cdot h_j), &  y_{PI}(i,j)=0
%         \end{array}\right.
%         \label{eq:mask}
% \end{equation}
% where $y_{PI}(i,j)$ denotes the PI label between node $i$ and $j$ and $\sigma(\cdot)$ is the sigmoid function. The confidence map $M$ works because \emph{it measures the predictive confidence by looking at the closeness between the prediction and the given PI label.} If the prediction becomes close to the given labels more easily, then the reliability of the PI labels is higher and more attention should be paid for such node pairs. 

% Guided by such intuition, we introduce a re-weighting mechanism for the PI loss $ \ell_{PI} = \ell_{PI} \otimes M$, where the re-weighted PI loss is obtained by multiplying the confidence mask and its original PI loss in an element-wise way. However, since the GNN is trained with the noisy node class labels $\overline{y}$ at the same time, \emph{the confidence mask $M$ cannot be estimated well if we entangle both the node label prediction and confidence estimation in a single GNN}. 

\textbf{Decoupling with two branches.} In this paper, as shown in Figure~\ref{fig:overview}, we propose to decouple the PI label estimation and node classification by using two separate GNNs, which are referred as a PI label estimation model $f_e$ and a node classification model $f_t$. The PI label estimation model generates the predictive confidence $y_{\text{PI}}$ by only learning with the PI estimation objective (Equation~\ref{eq:pi_loss}). The node classification model uses the generated PI label $y_{\text{PI}}$ from the PI label estimation model at the same time for model regularization.
% \vspace{-1em}
% \begin{equation}
%     \ell_{PI}^t = \ell_{PI}^t \otimes M,
%     \label{eq:uncertainty}
%     % \vspace{-0.7em}
% \end{equation}
% where $\ell_{PI}^t$ denotes the PI loss for the task executor $f_t$ and $\otimes$ means element-wise multiplication.

\textbf{Overall training.} Put them together, we introduce a new noise-robust training objective for node classification against noisy labels on GNNs, leveraging the estimated PI label in Section~\ref{sec:pi_gnn}. The key idea is to perform the node classification task by the node classification model $f_t$ while regularizing $f_t$ to produce similar embeddings for nodes that have a larger PI label and vice versa. Different from Equation~\ref{eq:pi_loss} that uses the discrete labels of 0 and 1, we use the smoothed PI label as the learning target. The overall noise-robust training objective for the node classification branch $f_t$ is formulated as:
\begin{equation}
    \mathcal{L}_{t} = \mathcal{L}_{\text{cls}}(f_t(A,X,\theta_t), \overline{y}) + \beta \cdot \mathcal{L}_{\text{reg}},
    \label{eq:all}
\end{equation}

where $\beta$ is a hyperparameter to balance the node classification loss $\mathcal{L}_{\text{cls}}$ and the regularization loss $\mathcal{L}_{\text{reg}}$ using the estimated PI labels. \textcolor{black}{Concretely, $\mathcal{L}_{\text{reg}}$ is defined as follows:}
\begin{equation}
    % \color{blue}
    \begin{aligned}
    \mathcal{L}_{\text{reg}} &= \lambda \cdot \mathbb{E}_{{(i,j) \in B^{-}}}\left[-(1-y_{\text{PI}}(i,j)) \cdot\log \frac{1}{1+\exp ^{h_i^T\cdot h_j}}-y_{\text{PI}}(i,j)\cdot \log \frac{\exp ^{h_i^T\cdot h_j}}{1+\exp ^{h_i^T\cdot h_j}}\right]\\
    & + \mathbb{E}_{{(i,j) \in B^{+}}}\left[-(1-y_{\text{PI}}(i,j)) \cdot\log \frac{1}{1+\exp ^{h_i^T\cdot h_j}}-y_{\text{PI}}(i,j)\cdot \log \frac{\exp ^{h_i^T\cdot h_j}}{1+\exp ^{h_i^T\cdot h_j}}\right]\\
     \end{aligned}
\end{equation}
\textcolor{black}{where $y_{\text{PI}}(i,j)$ is defined in Equation~\ref{eq:estimate}.}

Besides, the PI label estimation model is trained only by the binary classification loss $\mathcal{L}_e$ (Equation~\ref{eq:pi_loss}) and provides the estimated PI labels for model regularization on the node classification model, which is not affected by the noisy class labels. During inference, we discard the PI label estimation model $f_e$ and only use the node classification model $f_t$ for evaluation, which does not affect the inference speed.

Practically, the learning procedure relies heavily on the quality of the PI label estimation by $f_e$. Therefore, in the implementation, we pretrain the PI label estimation model $f_e$ for $K$ epochs (meanwhile we set $\beta$ in Equation~\ref{eq:all} to 0 to train the node classification model as well) and then jointly train the two models together by Equations~\ref{eq:pi_loss} and~\ref{eq:all}, respectively. Our algorithm is summarized in Algorithm~\ref{alg:algo}. 
% Note that the task executor is still optimized by the unweighted PI loss and the noisy classification loss during pretraining $f_m$. 

\textbf{Time complexity analysis.} Assume the GNN architecture for the node classification model and the PI label estimation model is GCN, for the node classification model, the time complexity is $\mathcal{O}\left(T L|E| d+T L |V| d^{2}\right)$ according to~\cite{wang2021decoupled} where $d$ is the dimension of the node embedding, $|E|,|V|$ are the number of edges and nodes of the graph and $T,L$ are the number of iterations and the layers. For the PI label estimation model, the complexity is $\mathcal{O}\left(T L|E| d+T L |V| d^{2} + T |V|^2d^2\right)$. For big graph datasets with large $|E|$ and $|V|$, we perform subgraph sampling~\cite{DBLP:conf/nips/HamiltonYL17} in the implementation (Section~\ref{sec:different_datasets}) to reduce the time complexity. 

\textcolor{black}{\textbf{Underlying assumption and limitation.} We note that  the proposed method \model has a limitation during deployment, which relies on the assumption about the underlying graph structure and their label distribution. Namely, homophilous graphs (\emph{i.e}., if two nodes are connected, then with high probability, they should have the same node label) are more suited to \model compared to  heterophilous graphs~\citep{zhu2020beyond,lim2021large,chien2022node}. As a quick verification, we test the performance of \model on several heterophilous datasets in Appendix Section~\ref{sec:app_het}, where the improvement of \model is less obvious compared to \model on homophilous graphs.}

% \vspace{-1em}

% \vspace{-2em}
%\emph{We present the sensitivity analysis results on the pretraining epochs $K$ and the regularization loss weight $\beta$ in Appendix Section~\ref{sec:app_sen}.}

\section{Experiments and Results}
% \vspace{-1em}
In this section, we present empirical evidence to validate the effectiveness of \model on different datasets with different noise types and ratios.
% \vspace{-1em}
\subsection{Experimental setting}
\label{sec:setting}

\begin{table*}[!t]
    \centering
    \tabcolsep 0.04in\renewcommand\arraystretch{0.745}{\small{}}%
    \scriptsize
    % \vspace{-0.5em}
     
    \resizebox{0.99\textwidth}{!}{%
    \begin{tabular}{cccccccccc}
\hline
\multicolumn{1}{c|}{Noise type}    & \multicolumn{1}{c|}{\cellcolor[HTML]{ECF4FF}No Noise}             & \multicolumn{4}{c|}{Symmetric Noise}                                                                           & \multicolumn{4}{c}{Asymmetric Noise}                                                     \\ \hline
\multicolumn{10}{c}{Cora}                                                                                                                                                                                                                                                                                           \\ \hline
\multicolumn{1}{c|}{Noise ratio}   & \multicolumn{1}{c|}{\cellcolor[HTML]{ECF4FF}0.0}                  & 0.2                  & 0.4                  & 0.6                  & \multicolumn{1}{c|}{0.8}                  & 0.2                  & 0.4                  & 0.6                  & 0.8                  \\ \hline
\multicolumn{1}{c|}{GCN}           & \multicolumn{1}{c|}{\cellcolor[HTML]{ECF4FF}\textbf{0.804(0.01)}} & 0.722(0.03)          & 0.613(0.07)          & 0.446(0.06)          & \multicolumn{1}{c|}{0.285(0.07)}          & 0.703(0.04)          & 0.514(0.06)          & 0.291(0.04)          & 0.161(0.02)          \\
\multicolumn{1}{c|}{\model w/o pc} & \multicolumn{1}{c|}{\cellcolor[HTML]{ECF4FF}0.781(0.01)}          & 0.731(0.02) & 0.654(0.05)          & 0.510(0.04)          & \multicolumn{1}{c|}{0.287(0.06)}          & 0.717(0.04)          & 0.563(0.07)          & 0.332(0.06) & 0.209(0.06) \\
\multicolumn{1}{c|}{\model}        & \multicolumn{1}{c|}{\cellcolor[HTML]{ECF4FF}0.780(0.01)}          & \textbf{0.739(0.02)}          & \textbf{0.664(0.03)} & \textbf{0.515(0.03)} & \multicolumn{1}{c|}{\textbf{0.296(0.05)}} & \textbf{0.723(0.03)} & \textbf{0.587(0.07)} & \textbf{0.350(0.07)}          & \textbf{0.232(0.06)}          \\ \hline
\multicolumn{10}{c}{CiteSeer}                                                                                                                                                                                                                                                                                       \\ \hline
\multicolumn{1}{c|}{GCN}           & \multicolumn{1}{c|}{\cellcolor[HTML]{ECF4FF}0.683(0.01)}          & 0.603(0.02)          & 0.524(0.04)          & 0.382(0.04)          & \multicolumn{1}{c|}{0.230(0.03)}          & 0.595(0.03)          & 0.465(0.05)          & 0.281(0.05)          & 0.171(0.05)          \\
\multicolumn{1}{c|}{\model w/o pc} & \multicolumn{1}{c|}{\cellcolor[HTML]{ECF4FF}0.656(0.03)}          & 0.606(0.03)          & 0.526(0.05)          & 0.378(0.05)          & \multicolumn{1}{c|}{0.227(0.04)}          & 0.588(0.04)          & 0.472(0.05)          & 0.328(0.03)          & 0.235(0.03)          \\
\multicolumn{1}{c|}{\model}        & \multicolumn{1}{c|}{\cellcolor[HTML]{ECF4FF}\textbf{0.684(0.03)}} & \textbf{0.642(0.03)} & \textbf{0.591(0.03)} & \textbf{0.432(0.07)} & \multicolumn{1}{c|}{\textbf{0.245(0.05)}} & \textbf{0.628(0.03)} & \textbf{0.531(0.06)} & \textbf{0.353(0.06)} & \textbf{0.238(0.06)} \\ \hline
\multicolumn{10}{c}{PubMed}                                                                                                                                                                                                                                                                                         \\ \hline
\multicolumn{1}{c|}{GCN}           & \multicolumn{1}{c|}{\cellcolor[HTML]{ECF4FF}\textbf{0.786(0.01)}} & 0.707(0.02)          & 0.610(0.06)          & 0.462(0.07)          & \multicolumn{1}{c|}{0.367(0.07)}          & 0.682(0.05)          & 0.524(0.08)          & 0.399(0.06)          & 0.387(0.07)          \\
\multicolumn{1}{c|}{\model w/o pc} & \multicolumn{1}{c|}{\cellcolor[HTML]{ECF4FF}0.774(0.00)}          & 0.723(0.03)          & 0.628(0.05)          & 0.458(0.07)          & \multicolumn{1}{c|}{0.370(0.06)} & 0.722(0.03)          & 0.579(0.07)          & 0.412(0.05)          & 0.401(0.03) \\
\multicolumn{1}{c|}{\model}        & \multicolumn{1}{c|}{\cellcolor[HTML]{ECF4FF}0.774(0.00)}          & \textbf{0.724(0.03)} & \textbf{0.638(0.04)} & \textbf{0.470(0.08)} & \multicolumn{1}{c|}{\textbf{0.379(0.07)}}          & \textbf{0.723(0.03)} & \textbf{0.583(0.07)} & \textbf{0.425(0.07)} & \textbf{0.406(0.04)}          \\ \hline
\multicolumn{10}{c}{WikiCS}  \\ \hline
\multicolumn{1}{c|}{GCN}           & \multicolumn{1}{c|}{\cellcolor[HTML]{ECF4FF}\textbf{0.703(0.01)}} & 0.635(0.03)          & 0.558(0.04)          & 0.376(0.05)          & \multicolumn{1}{c|}{0.183(0.05)}          & 0.608(0.05)          & 0.468(0.05)          & 0.272(0.05)          & 0.129(0.07)          \\
\multicolumn{1}{c|}{\model w/o pc} & \multicolumn{1}{c|}{\cellcolor[HTML]{ECF4FF}0.676(0.01)}          & 0.624(0.02)          & 0.552(0.05)          & 0.396(0.07)          & \multicolumn{1}{c|}{0.197(0.07)}          & 0.607(0.03)          & 0.470(0.05) & 0.290(0.05) & 0.125(0.05)          \\
\multicolumn{1}{c|}{\model}        & \multicolumn{1}{c|}{\cellcolor[HTML]{ECF4FF}0.676(0.01)}          & \textbf{0.636(0.02)} & \textbf{0.562(0.04)} & \textbf{0.398(0.07)} & \multicolumn{1}{c|}{\textbf{0.208(0.07)}} & \textbf{0.610}(0.04) & \textbf{0.483(0.05)}          & \textbf{0.303(0.04)}          & \textbf{0.135(0.06)} \\ \hline

% \multicolumn{10}{c}{DBLP}                                                                                                                                                                                                                                                                                           \\ \hline
% \multicolumn{1}{c|}{GCN}           & \multicolumn{1}{c|}{\cellcolor[HTML]{ECF4FF}\textbf{0.641(0.02)}} & 0.542(0.09)          & 0.448(0.08)          & 0.266(0.04)          & \multicolumn{1}{c|}{0.246(0.06)}          & 0.503(0.10)          & 0.376(0.08)          & 0.284(0.09)          & 0.204(0.08)          \\
% \multicolumn{1}{c|}{CPL w/o ce} & \multicolumn{1}{c|}{\cellcolor[HTML]{ECF4FF}0.622(0.05)}          & 0.560(0.12) & 0.455(0.12)          & 0.294(0.08)          & \multicolumn{1}{c|}{0.253(0.09)}          & 0.521(0.08)          & 0.399(0.09)          & 0.327(0.09) & 0.260(0.12) \\
% \multicolumn{1}{c|}{CPL}        & \multicolumn{1}{c|}{\cellcolor[HTML]{ECF4FF}0.635(0.04)}          & \textbf{0.566(0.13)}          & \textbf{0.456(0.10)} & \textbf{0.301(0.08)} & \multicolumn{1}{c|}{\textbf{0.258(0.11)}} & \textbf{0.558(0.08)} & \textbf{0.453(0.09)} & \textbf{0.335(0.11)}          & \textbf{0.292(0.14)}          \\ \hline
\multicolumn{10}{c}{OGB-arxiv}                 \\ \hline                                                                                                                                                                                            \multicolumn{1}{c|}{GCN}           &\multicolumn{1}{c|}{\cellcolor[HTML]{ECF4FF}\textbf{0.491(0.01)}} & 0.461(0.01) & 0.433(0.01) & 0.393(0.03) & \multicolumn{1}{c|}{0.278(0.02)} & 0.435(0.01) & 0.399(0.01) & 0.059(0.01) & \multicolumn{1}{l}{0.019(0.00)}       \\
\multicolumn{1}{c|}{\model w/o pc} &  \multicolumn{1}{c|}{\cellcolor[HTML]{ECF4FF} 0.462(0.04)} & 0.469(0.08) & 0.445(0.05) & 0.406(0.05) & \multicolumn{1}{c|}{0.357(0.10) }& 0.445(0.08) & 0.425(0.06) & 0.060(0.02) & \multicolumn{1}{l}{0.021(0.00)}     \\
\multicolumn{1}{c|}{\model}        & \multicolumn{1}{c|}{\cellcolor[HTML]{ECF4FF}0.482(0.01)} & \textbf{0.476(0.04)} & \textbf{0.467(0.03)} & \textbf{0.418(0.04)} &\multicolumn{1}{c|}{\textbf{0.368(0.09)}} & \textbf{0.475(0.01)} & \textbf{0.461(0.01)} & \textbf{0.069(0.01)} & \textbf{0.022(0.00)}  \\ \hline                                                       

\end{tabular}
}
\caption{\small Test accuracy on 5 datasets for \model with GCN as the backbone. \textbf{Bold} numbers are superior results. Std. is shown in the bracket. w/o pc means that the PI labels are not estimated using the predictive confidence but just the node connectivity.}
% \vspace{-4ex}
    \label{tab:effectiveness}
\end{table*}

\textbf{Datasets.} We used five datasets to evaluate \model, including Cora, CiteSeer and PubMed with the default dataset split as in~\citep{DBLP:conf/iclr/KipfW17} and WikiCS dataset~\citep{DBLP:journals/corr/abs-2007-02901} as well as 
OGB-arxiv dataset~\citep{DBLP:conf/nips/HuFZDRLCL20}.
% .
% DBLP~\citep{DBLP:conf/ijcai/PanWZZW16}
For WikiCS, we used the first 20 nodes from each class for training and the next 20 nodes for validation. The remaining nodes for each class are used as the test set. For OGB-arxiv, we use the default split. \emph{The dataset statistics are summarized in Appendix Section~\ref{sec:app_dataset}.}

% \end{table}
Since all datasets are clean, following~\cite{DBLP:conf/cvpr/PatriniRMNQ17}, we corrupted these datasets manually by the
noise transition matrix $Q_{i j}=\operatorname{Pr}(\overline{y}=j \mid y=i)$ given that noisy $\overline{y}$ is flipped from clean $y$. Assume the matrix $Q$ has two representative structures: (1) Symmetry flipping~\citep{DBLP:conf/nips/RooyenMW15}; (2) Asymmetric pair flipping: a simulation of fine-grained classification with noisy labels, where labelers may
make mistakes only within very similar classes. Note the asymmetric case is much harder than the symmetry case. \emph{Their precise definition is in Appendix Section~\ref{sec:app_def}.} We tested four different noise rates $\epsilon\in\{0.2,0.4,0.6,0.8\}$ in this paper for two different noise types, which cover lightly and extremely noisy supervision. Note that in the most extreme case, the noise rate $80\%$ for pair flipping means $80\%$ training data have wrong labels that cannot be learned without additional assumptions.

\textbf{Implementation details.} We used three different GNN architectures, \emph{i.e.}, GCN, GAT and GraphSAGE, which are implemented by~\url{torch-geometric}~\footnote{\url{https://github.com/pyg-team/pytorch_geometric/blob/master/examples}}~\citep{Fey/Lenssen/2019}. All of them have two layers. Specifically, the hidden dimension of GCN, GAT and GraphSAGE is set to 16, 8 and 64. GAT has 8 attention heads in the first layer and 1 head in the second layer. The mean aggregator is used for GraphSAGE. We applied Adam optimizer~\citep{DBLP:journals/corr/KingmaB14} with a learning rate of 0.01 for GCN and GraphSAGE and 0.005 for GAT. The weight decay is set to 5e-4. We trained for 400 epochs on a Tesla P40. The loss weight $\beta$ is set to $|V|^2/(|V|^2-Q)^2$, where $|V|$ is the number of nodes and $Q$ is the sum of all elements of the preprocessed adjacency matrix. The number of pretraining epochs $K$ is set to 50 and the total epoch $N$ is set to 400. \textcolor{black}{For subgraph sampling, we sampled 15 and 10 neighbors for each node in the 1st and 2nd layer of the GNN and set the batch size to be 1024.} We tuned all the hyperparameters on the validation set and reported the node classification accuracy on the clean test set. \emph{Details about the ablation studies on these factors are shown in Section~\ref{sec:ablation}}. Each experiment is repeated for 10 times with random seeds from 1 to 10. \emph{The training time for our \model and the vanilla GNN  model is compared in the Appendix Section~\ref{sec:training_time}}.
% \vspace{-1em}
%The PI label is constructed based on the adjacency matrix by default

\begin{figure}[t]
% \vspace{-1.5em}
  \begin{center}
    \includegraphics[width=1.0\textwidth]{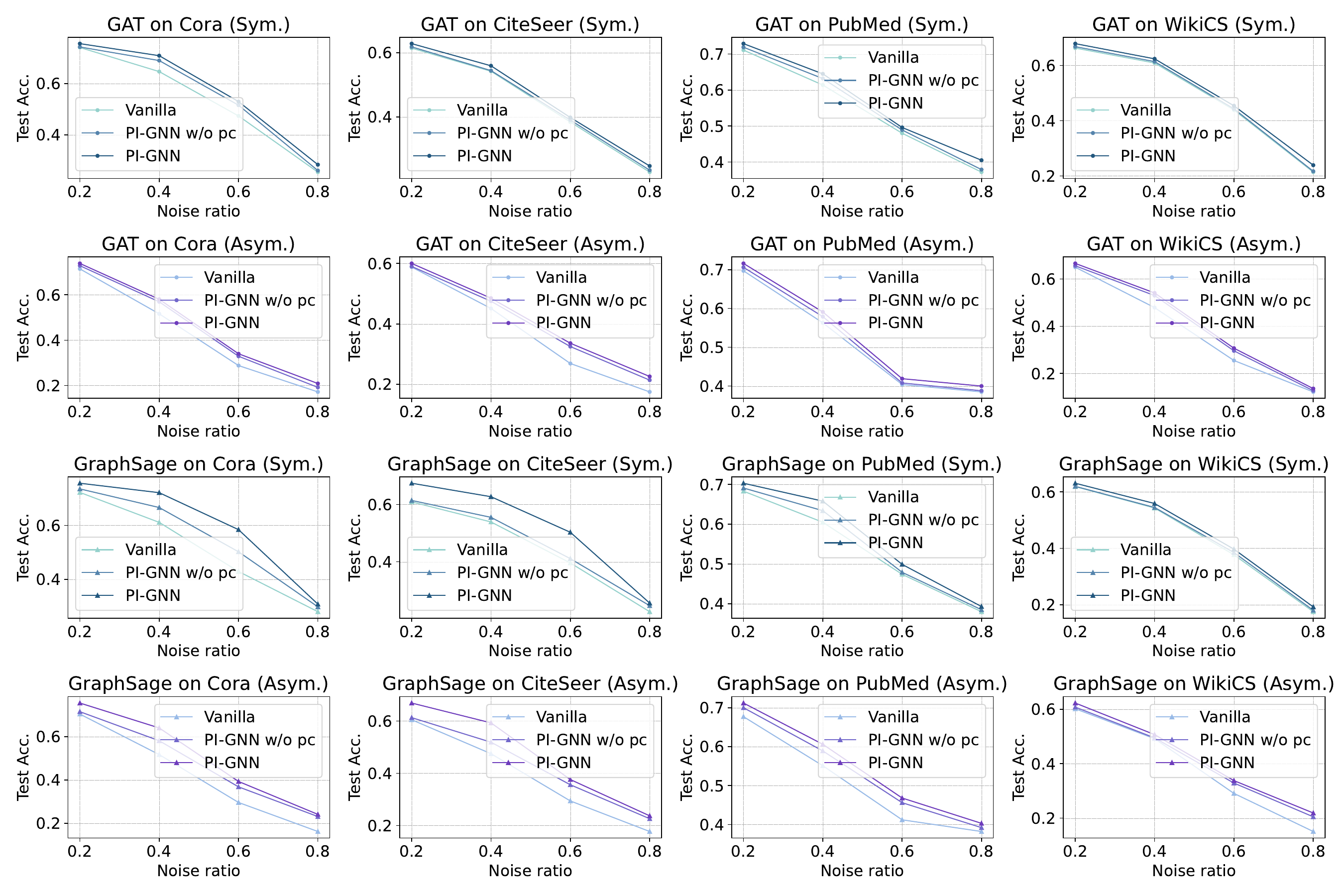}
  \end{center}
  % \vspace{-1.2em}
 \caption{\small Test accuracy of \model and comparison with \model w/o pc and vanilla GNN on two additional model architectures under different noisy settings. }
  % \vspace{-1.5em}
  \label{fig:architectures}
\end{figure}

\subsection{Effectiveness on different datasets}
\label{sec:different_datasets}
We evaluated the effectiveness of \model on five datasets with different noisy labels and noise rates, which is shown in Table~\ref{tab:effectiveness} with GCN as the backbone. Specifically, we are interested to observe 1) whether the introduced regularization objective between nodes can improve a vanilla GNN against noisy labels and 2) whether the predictive confidence from the PI label estimation model is beneficial for the test accuracy. Therefore, we compared the accuracy of a vanilla GNN, \model trained with the node connectivity as the PI label (\model w/o pc) and \model.

From Table~\ref{tab:effectiveness}, we made several observations: \textbf{Firstly}, the GNN trained with the PI regularization objective is more robust to noisy labels, where both \model w/o pc and \model perform much better than a vanilla GNN. \textbf{Secondly}, by estimating the PI labels using the PI label estimation model, the test accuracy is further improved compared to directly using the node connectivity as the PI label. For instance, \model improves the accuracy by 2.3\% with the asymmetric noise (noise ratio $\epsilon=0.8$) on Cora compared to \model w/o pc, which justifies the effectiveness of our design. \textbf{Thirdly}, the \model does not help the GNN with the clean node labels, \emph{e.g.}, 80.4\% of a vanilla GCN vs. 78.0\% of \model on Cora, which illustrates the \model helps to combat noisy supervision rather than inherently improve the node classification with purely clean node labels. \emph{Additional results on heterophilous datasets and with lower noise ratios are in Appendix Sections~\ref{sec:app_het} and~\ref{sec:app_small_ratio}.}

\subsection{Performance on different GNN architectures}
We evaluated \model on different GNN architectures, \emph{i.e.}, GAT and GraphSAGE. The experiments are conducted on Cora, CiteSeer, PubMed and WikiCS datasets, which are shown in Figure~\ref{fig:architectures}. As can be observed, \model performs similarly on GAT and GraphSAGE compared to the results on GCN, where the regularization of PI and the  predictive confidence are both beneficial for model generalization even with extremely noisy supervision. Moreover, using the predictive confidence as PI labels is more effective on GraphSAGE. For example, in the Cora dataset, \model improves \model w/o pc by 4.2\% and 3.1\% on average under symmetric and asymmetric noise, respectively, which is larger than that for GAT and GCN. It may suggest the mean aggregator in GraphSAGE is more susceptible to the sub-optimal PI labels. \emph{We provide significance test on the results of GAT in Appendix Section~\ref{sec:app_sig}.}

% Table generated by Excel2LaTeX from sheet 'Cora'
\begin{table*}[t]
% \tiny
\scriptsize
  \centering

  \tabcolsep 0.04in\renewcommand\arraystretch{0.9}{\small{}}%
  
  % \fontsize{7.8pt}{7.8pt}\selectfont
   \scalebox{1.0}{ 
    \begin{tabular}{c|cc|cc}
    % \hline
    Noise type & \multicolumn{2}{c}{Symmetric Noise}   & \multicolumn{2}{c}{Asymmetric Noise} \\
    \hline
    Noise ratio &  0.4   & 0.6     & 0.4 & 0.6  \\
   \hline
   & \multicolumn{4}{c}{Test dataset: Cora / CiteSeer} \\

 \hline
  Decoupling&0.581(0.06) / 0.518(0.03) & 0.425(0.06) / 0.390(0.03) & 0.541(0.05) / 0.474(0.04) & 0.336(0.03) / 0.323(0.05)\\
     GCE&0.627(0.07) / 0.530(0.03) & 0.447(0.06) / 0.383(0.03) & 0.511(0.05) / 0.468(0.05)& 0.284(0.03) / 0.285(0.05) \\
    APL & 0.624(0.08) / 0.522(0.04) & 0.446(0.06) / 0.376(0.04) &  0.507(0.06) / 0.456(0.06) & 0.281(0.03) / 0.281(0.04)  \\
   Co-teaching& 0.577(0.11) / 0.573(0.07) & 0.376(0.07) / 0.404(0.06)  & 0.457(0.10) / 0.462(0.08) &0.237(0.09) / 0.256(0.08)\\
     LPM-1&0.542(0.09) / 0.467(0.06) & 0.447(0.07) / 0.395(0.08) &0.481(0.07) / 0.506(0.08) & 0.318(0.04) / 0.341(0.09)\\
   T-Revision&  0.596(0.06) / 0.518(0.03) & 0.425(0.06) / 0.380(0.04) &0.512(0.06) / 0.457(0.06) & 0.281(0.05) / 0.263(0.05)  \\
   DivideMix& 0.628(0.06) / 0.515(0.05) & 0.463(0.09) / 0.355(0.05) &  0.428(0.01) / 0.396(0.03) & 0.313(0.03) / 0.282(0.02)\\
   \hline
  \rowcolor{Gray}    \model (ours)& \textbf{0.664(0.03) / 0.591(0.03)} & \textbf{0.515(0.03) / 0.432(0.07)} &  \textbf{0.587(0.07) / 0.531(0.06)} & \textbf{0.350(0.07)} / \textbf{0.353(0.06)}\\
    
    % \hline 
 \end{tabular}}
 \caption{\small Comparative results with baselines. \textbf{Bold} numbers are superior results. LPM-1 means one extra clean label is used for each class. The result on the left and right of each cell is the classification accuracy of the Cora dataset and CiteSeer dataset, respectively. }
  \label{tab:baselines}%
  % \vspace{-1.5em}
\end{table*}%
% \vspace{-1em}
\subsection{Comparison with baselines}
\label{sec:baseline}
% \vspace{-1em}
In order to further demonstrate the competitive performance of \model, we compared with several powerful baselines for combating noisy labels in literature. For a fair comparison, we used the same GNN architecture, \emph{i.e.}, GCN, and the same overlapping hyperparameters during implementation. The other method-specific hyperparameters are tuned according to the original paper on the validation set. Specifically, we compared with noise-transition matrix-based method, T-revision~\citep{DBLP:conf/nips/XiaLW00NS19}, robust loss functions, such as Generalized Cross Entropy (GCE) loss~\citep{DBLP:conf/nips/ZhangS18} and Active Passive Loss (APL)~\citep{DBLP:conf/icml/MaH00E020}, optimization-based approaches, such as Co-teaching~\citep{DBLP:conf/nips/HanYYNXHTS18}, Decoupling~\citep{DBLP:conf/nips/MalachS17} and DivideMix~\citep{DBLP:conf/iclr/LiSH20}. We also compared with Label Propagation and Meta learning (LPM)~\citep{xia2021towards}, a method that is specifically designed for solving label noise for node classification but uses a small set of clean labels. For illustration, we reported the classification accuracy on Cora and CiteSeer with symmetric and asymmetric noise (noise rate $\epsilon=0.4,0.6$) in Table~\ref{tab:baselines}.

From Table~\ref{tab:baselines}, \model outperforms different baselines with a considerable margin, \emph{e.g.}, improving the classification accuracy by 5.2\% on Cora under the symmetric noise ($\epsilon=0.6$) compared to the best baseline. Moreover, \model is able to outperform LPM-1, which relieves the strong assumption that auxiliary clean node labels are available. \emph{We compare with traditional graph semi-supervised learning approaches and on more datasets in Appendix Section~\ref{sec:app_trad} and Section~\ref{sec:baselines_appendix}, respectively.}
% \vspace{-1em}

% \vspace{-1em}
\subsection{Ablation studies}

\textbf{The effect of PI labels.} To show the importance of informative PI labels, we tested \model under different PI labels in addition to the one that is estimated by the PI label estimation model (ours): 1) random labels, 2) noisy class label comparison and 3) estimated PI w/o $f_e$  where only one GNN is used for both node classification and PI label estimation. We also compared with the PI labels by clean class label comparison, which is expected to be the oracle. We used GCN as the backbone and tested on two datasets, \emph{i.e.}, Cora and CiteSeer and different noise types. The results are shown in Table~\ref{tab:different_pi_label}.
\begin{wraptable}{r}{0.45\linewidth}
\centering
% \vspace{2pt}
% \vspace{-10pt}
\resizebox*{\linewidth}{!}{%
\begin{tabular}{ccc}
\toprule
Noise Type & Symmetric Noise & Asymmetric Noise \\ \midrule
\multicolumn{3}{c}{Cora} \\
\hline
Noise Ratio &  0.6 & 0.6\\
\hline

% Clean label comparison & 0.453(0.05)&0.300(0.05)\\
Random PI label &0.449(0.05)&0.295(0.04)  \\ 
Noisy label comparison & 0.471(0.05)& 0.300(0.04) \\
Estimated PI w/o $f_e$& 0.511(0.03) & 0.329(0.05)\\
\textcolor{gray}{Clean label comparison (oracle)} &\textcolor{gray}{0.447(0.07)} & \textcolor{gray}{0.355(0.05)} \\ 
\cellcolor{Gray} Estimated PI (ours) & \cellcolor{Gray}\textbf{0.515(0.03)}& \cellcolor{Gray}\textbf{0.350(0.07)}\\

\bottomrule
\multicolumn{3}{c}{CiteSeer} \\
\hline
Noise Ratio & 0.6& 0.6 \\
\hline

% Clean label comparison &0.341(0.05)& 0.283(0.04) \\
Random PI label & 0.351(0.04)& 0.298(0.04) \\ 
Noisy label comparison &0.378(0.04)&0.312(0.04)  \\
Estimated PI w/o $f_e$ & 0.430(0.07) &0.340(0.05) \\
\textcolor{gray}{Clean label comparison (oracle)} &\textcolor{gray}{0.447(0.07)} & \textcolor{gray}{0.355(0.05)} \\ 
\cellcolor{Gray}Estimated PI (ours) & \cellcolor{Gray}\textbf{0.432(0.07)}&\cellcolor{Gray}\textbf{0.353(0.06)}\\

\hline
\end{tabular}}
% \vspace{-1em}
\caption{\small Performance of \model with different PI labels.}
\label{tab:different_pi_label}
% \vspace{-15pt}
\end{wraptable}

From Table~\ref{tab:different_pi_label}, using PI labels based on the PI label estimation model achieves better performance than using noisy node label comparison. Employing randomly generated PI labels incurs the worst performance because it completely ignores the importance of the pairwise interactions.~Moreover, removing the PI label estimation model decreases the test accuracy because the node embeddings are optimized by the noisy class labels as well, which might not be effective for the PI label estimation (32.9\% vs.~35.0\% for Cora with 60\% asymmetric noise). Finally, our \model achieves a similar performance compared to training by clean label comparison (the oracle case), showcasing the effectiveness of our estimated PI labels. Note in our noisy setting, it is impossible to reach the oracle. \emph{More results are in Appendix Section~\ref{sec:app_clean}.}

% \emph{We also compare with training with the clean PI labels, which can be regarded as an upper bound of the proposed \model in Appendix Section~\ref{sec:app_clean}.}

% \section{Sensitivity analysis}
% \label{sec:app_sen}

% \vspace{-1em}
\begin{figure*}[t]
    \centering
    \includegraphics[width=\textwidth]{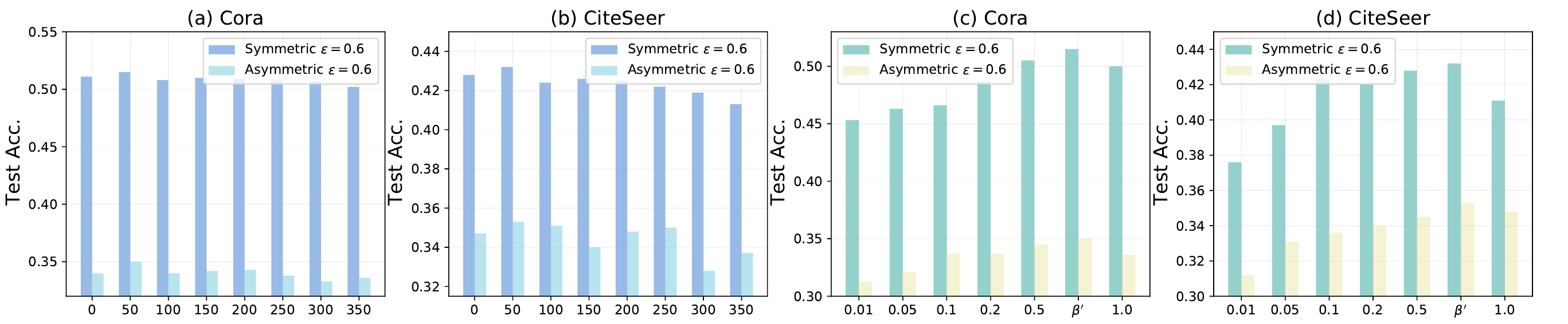}
    % \vspace{-1.5em}
    \caption{ \small(a)-(b) Performance of \model \emph{w.r.t.} different pretraining epochs on Cora and CiteSeer. x axis denotes the value of the pretraining epochs for the PI label estimation model. (c)-(d) Performance of \model \emph{w.r.t.} the regularization loss weight $\beta$. x axis denotes the value of the loss weight and $\beta^{\prime}$ is the weight that is aware of the sparsity of the input graph (\emph{cf.} Section~\ref{sec:setting}).}%while $\scriptsize{f_{W_i^j}(\cdot)}$ means the atomic block with parameter $W_i^j$
    \label{fig:ablation_1}
    % \vspace{-2em}
\end{figure*}

\textbf{Sensitivity to the pretraining epoch of the PI label estimation model.} We investigated whether the performance of \model is sensitive to the number of pretraining epochs for the PI label estimation model. The experimental results on Cora and CiteSeer with GCN under symmetric and asymmetric noise are shown in Figure~\ref{fig:ablation_1}~(a) and (b). For illustration, we set the noise ratio to be 0.6. As can be observed, pretraining the PI label estimation model for $K$ epochs is effective for improving the generalization on the clean test set. Given a small $K$, the confidence mask is not estimated well which is not helpful to apply it on the node classification model for regularization. Meanwhile, $K$ should not be too large in order to sufficiently regularize the node classification model using Equation~\ref{eq:all}. $K$ is set to 50 for all the experiments.

\begin{wrapfigure}{r}{0.65\textwidth}
\vspace{-1.5em}
  \begin{center}
    \includegraphics[width=0.65\textwidth]{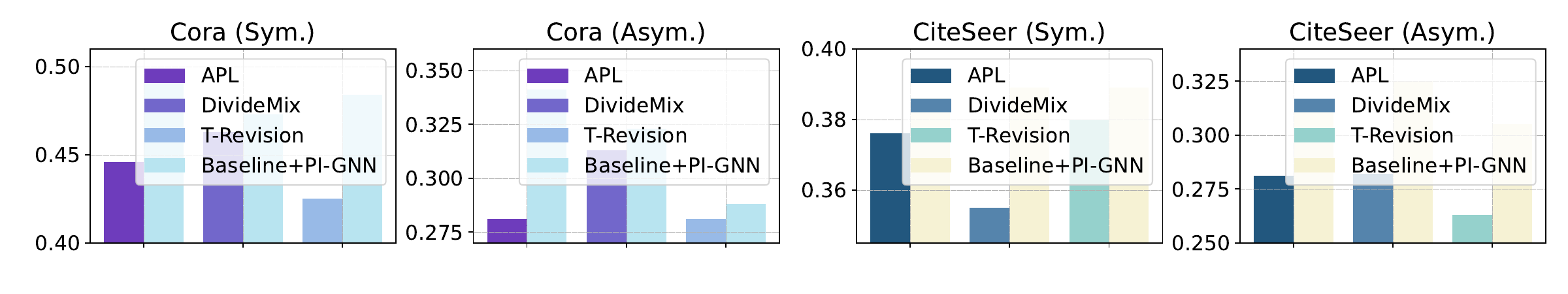}
  \end{center}
  % \vspace{-1.2em}
 \caption{\small Performance of the \model applied on different label-noise baselines on Cora and CiteSeer. Noise ratio is set to 0.6. }
  % \vspace{-1.5em}
  \label{fig:add_baseline}
\end{wrapfigure}

\textbf{Application of \model on label-noise baselines.} To observe whether \model is able to improve the generalization ability for different label-noise baseline models, we extended three representative approaches, \emph{i.e.}, T-revision~\citep{DBLP:conf/nips/XiaLW00NS19}, APL~\citep{DBLP:conf/icml/MaH00E020} and DivideMix~\citep{DBLP:conf/iclr/LiSH20} by adding the PI regularization objective during training. Specifically, we used the sum of their original loss and the PI regularization loss ($\mathcal{L}_{\text{reg}}$, \emph{cf.} Equation~\ref{eq:all}) to optimize the GNN. The weight for PI regularization loss is set to the same value as our approach (\emph{cf.} Section~\ref{sec:setting}). We chose GCN as the backbone and reported the test accuracy on Cora and CiteSeer with both symmetric and asymmetric noise ($\epsilon=0.6$) in Figure~\ref{fig:add_baseline}. As the result shows, \model is orthogonal to those noise-robust baseline models, which is potentially useful for improving their performance without bells and whistles. For instance, The test set accuracy is improved by 4.3\% on DivideMix under the asymmetric noise for the CiteSeer dataset and thus demonstrates the universality of our proposed \model.

% \begin{figure}[t]
%     \centering
%     \includegraphics[width=0.5\textwidth]{figs/add_baseline.pdf}
%     \vspace{-2.5em}
%     \caption{ \small Performance of the \model applied on different label-noise baselines on Cora and CiteSeer. Noise ratio is set to 0.6.}%while $\scriptsize{f_{W_i^j}(\cdot)}$ means the atomic block with parameter $W_i^j$
%     \label{fig:add_baseline}
%     \vspace{-2em}
% \end{figure}

\begin{wraptable}{r}{0.4\linewidth}
\centering
% \vspace{-20pt}
% \vspace{-10pt}

\resizebox*{\linewidth}{!}{%
    \begin{tabular}{c|cc}
    \hline
         Noise Type& Sym. Noise & Asym. Noise  \\
         \hline
         Noise Ratio& 0.6 & 0.6\\
         \hline
         GAT only & 0.394(0.05) & 0.330(0.04) \\
    GCN-GAT & \textbf{0.412(0.06)} & \textbf{0.339(0.03)} \\
    \hline
     GraphSAGE only & 0.503(0.08) & 0.376(0.08)\\
     GCN-GraphSAGE & \textbf{0.512(0.09)}&\textbf{ 0.383(0.06)}\\
    \hline
  GraphSAGE only & 0.503(0.08) & 0.376(0.08) \\
    GAT-GraphSAGE & \textbf{0.516(0.08)}& \textbf{ 0.381(0.06)}  \\
    \hline
    \end{tabular}
}
% \vspace{-0.5em}
\caption{\small Performance of \model with different architectures for two branches on CiteSeer.}
\label{tab:two_archi}
% \vspace{-11pt}
\end{wraptable}

\textbf{Different architectures for two branches.} \model allows for a flexible choice of the architectures for the PI label estimation model and the node classification model, where a light-weight PI label estimation model can help a large node classification model for node classification during  training. In what follows, we used three different PI label estimation-node classification model pairs, namely GCN-GAT, GCN-GraphSAGE and GAT-GraphSAGE. The number of parameters for GCN, GAT, GraphSAGE is 0.02, 0.09 and 0.18 M, respectively. The comparison with using the same architectures are shown in Table~\ref{tab:two_archi}. From Table~\ref{tab:two_archi}, using a light-weight GNN for the PI label estimation model is able to further improve the clean test accuracy, which is promising for efficient deployment of \model on real-world graph datasets.

\label{sec:ablation}
\textbf{The effect of regularization weight.} To observe whether the regularization loss weight $\beta$ matters to the model performance, we trained \model with different values of $\beta$, \emph{i.e.}, 0.01, 0.05, 0.1, 0.2, 0.5, 1.0 and compared with the value $\beta^{\prime}=|V|^2/(|V|^2-Q)^2$ which is aware of the sparsity of the graph in Figure~\ref{fig:ablation_1}~(c) and (d). We conducted experiments on Cora and CiteSeer with GCN and showed the results with symmetric and asymmetric noise ($\epsilon=0.6$). From the figure, \model is sensitive to the choice of regularization loss weight $\beta$. On both datasets with different noise types, \model trained with $\beta^{\prime}$ achieves the best test accuracy, and simultaneously avoids heavy tuning procedure on the validation set.

% \vspace{-1em}
\section{Related work}
% \vspace{-1em}
\textbf{Graph Neural Networks.} Graph neural networks have been widely used to model the graph-structured data with various architectures, such as graph convolutional network (GCN)~\citep{DBLP:conf/iclr/KipfW17}, graph attention network (GAT)~\citep{DBLP:conf/iclr/VelickovicCCRLB18}, GraphSAGE~\citep{DBLP:conf/nips/HamiltonYL17}, Graph Isomorphism Network (GIN)~\citep{DBLP:conf/iclr/XuHLJ19}, Simple Graph Convolution (SGC)~\citep{DBLP:conf/icml/WuSZFYW19}, etc. Common graph analysis tasks, including node classification~\citep{DBLP:conf/nips/LanWDSTG20}, link prediction~\citep{DBLP:conf/nips/ZhangC18}, graph classification~\citep{DBLP:conf/icml/BacciuEM18}, graph generation~\citep{DBLP:conf/nips/LiaoLSWHDUZ19,DBLP:conf/iclr/ShiXZZZT20}, have been widely studied in literature. However, only a few works focused on training robust GNNs against noisy labels, such as by loss correction~\citep{DBLP:journals/corr/abs-1905-01591} for graph classification, sample re-weighting~\citep{xia2021towards,DBLP:journals/corr/abs-2103-03414} for node classification. None of them exploited explicit PI, which are compared with our \model in Section~\ref{sec:baseline}.~\cite{DBLP:journals/corr/BuiRR17,DBLP:conf/nips/StretcuVMPRT19,DBLP:conf/nips/NgCS18,DBLP:conf/icml/QuBT19,ma2019cgnf} utilized graph structures for semi-supervised learning but with clean labels. Moreover, they did not further process the graph structure   while \model utilizes the graph structure and introduces a new PI label estimation procedure during training.~\cite{DBLP:conf/cvpr/JiangZLTL19,DBLP:conf/pkdd/YuZJWY20,DBLP:conf/icml/ZhengZCSNYC020,DBLP:conf/nips/0022WZ20,kim2021how,DBLP:journals/corr/abs-2102-05034} iteratively refined graph structure during training for missing edge prediction or error edge detection while \model does not change the input graph.\textcolor{black}{~\citet{luo2021learning} proposed a parameterized topological denoising network to improve the robustness and generalization performance of GNNs by learning to \textit{drop task-irrelevant edges}. The main difference is that \model deals with the situation where the node labels are corrupted while they deal with the noisy edges.}

\textcolor{black}{\citet{zhao2020uncertainty,stadler2021graph,wu2023energy} proposed to use uncertainty estimation and out-of-distribution detection techniques~\citep{du2022vos,du2022unknown,du2022siren,tao2023nonparametric,yang2021generalized,bai2023feed,sun2021react,sun2022dice,sun2022knn,ming2022delving,ming2022posterior,ming2023cider,wei2022mitigating,yang2022openood} for detecting out-of-distribution or noisy node samples on graphs, where our goal is to correctly classify the noisy nodes.}
%DBLP:conf/aaai/LiSCLTL20,DBLP:conf/www/LiRCMHH19,DBLP:conf/nips/YouLYPL18

\textbf{Neural networks with noisy labels.} Methods for neural networks against noisy labels can be roughly categorized into three types~\citep{DBLP:journals/corr/abs-2011-04406}, \emph{i.e.}, approaches from the perspective of data~\citep{DBLP:journals/jmlr/RooyenW17}, learning objective~\citep{DBLP:journals/corr/ReedLASER14,DBLP:journals/pami/MiyatoMKI19} and optimization~\citep{DBLP:conf/icml/ArpitJBKBKMFCBL17}. Methods based on data mainly 
built the noise transition matrix to explore
the data relationship between clean and noisy label by an adaptation layer~\citep{sukhbaatar2015training}, loss correction~\citep{DBLP:conf/cvpr/PatriniRMNQ17} and prior knowledge~\citep{DBLP:conf/nips/HanYNZTZS18}. Methods based on objective modified the learning objective by regularization~\citep{DBLP:conf/icml/00030YYXTS20}, reweighting~\citep{DBLP:journals/pami/LiuT16,DBLP:conf/icml/WangKB17} and loss redesign~\citep{DBLP:conf/icml/ThulasidasanBBC19}. Methods based on optimization mainly changed the optimization policy, such as by memorization effect~\citep{DBLP:conf/icml/JiangZLLF18}, self-training~\citep{DBLP:conf/icml/RenZYU18} and co-training~\citep{DBLP:conf/icml/Yu0YNTS19}. \cite{DBLP:journals/corr/abs-2006-07831} proposed to use the similarity loss for noisy labels of images but it relied on the noisy transition matrix, which is sensitive to the matrix estimation quality and cannot use the graph structure for regularization. In this paper, we extend several approaches from each category to compare with \model in Section~\ref{sec:baseline}.
%DBLP:conf/iclr/ZhangBHRV17,
%,DBLP:conf/iclr/GoldbergerB17
%DBLP:conf/nips/XiaLW00NS19
%DBLP:journals/corr/AzadiFJD15
%DBLP:conf/icml/CharoenphakdeeL19,
%DBLP:conf/nips/HanYYNXHTS18,
%,DBLP:conf/iclr/LiSH20
%DBLP:conf/nips/ZhangS18

% \vspace{-1em}
\section{Conclusion}
% \vspace{-1em}
In this paper, we proposed \model, a simple but effective learning paradigm for helping the GNN to generalize well with noisy supervision. Our key idea is to leverage the pairwise interactions between nodes to explicitly regularize the similarity of those node embeddings during training. In order to perform noise-robust node classification, we introduce a new learning framework to adaptively estimate and leverage the pairwise interactions for model regularization. We conducted
extensive experiments to demonstrate that \model can train GNNs robustly
under extremely noisy supervision, which serves as a crucial step towards the reliable deployment of GNNs in complex real-world applications. We
hope our work inspires future research on noise-robust graph learning, \textcolor{black}{such as proposing novel pairwise approaches from the algorithmic perspective and constructing real-world noisy graph datasets for a more comprehensive and practical empirical evaluation.}

\section{Acknowledgement}

BH was supported by Tencent AI Lab Rhino-Bird Gift Fund. The authors would also like to thank TMLR reviewers for the helpful suggestions and
feedback.

\bibliography{citations}
\bibliographystyle{tmlr}

\appendix
% \section{Appendix}

\onecolumn
\begin{center}
    \Large{\textbf{Noise-robust Graph Learning by Estimating and Leveraging
Pairwise Interactions
 \\ (Appendix)}}
\end{center}

% \section{Broad Impact}
% \label{sec:supp_impact}
%  Our project aims to improve the reliability and safety of modern graph neural networks against noisy node labels. Our study can lead to direct benefits and societal impacts, particularly for safety-critical applications such as trustworthy graph analysis in social networks and drug discovery. Our study does not involve any human subjects or violation of legal compliance. We do not anticipate any potentially harmful consequences to our work. Through our study and releasing our code, we hope to raise stronger research and societal awareness towards the problem of learning with noisy graphs in real-world settings. 

\section{Definition of noise}
\label{sec:app_def}
The definition of transition matrix $Q$ is as follows. $n$ is number of the class.

Asymmetric pair flipping:
\begin{equation}
    Q=\left[\begin{array}{ccccc}1-\epsilon & \epsilon & 0 & \ldots & 0 \\ 0 & 1-\epsilon & \epsilon & & 0 \\ \vdots & & \ddots & \ddots & \vdots \\ 0 & & & 1-\epsilon & \epsilon \\ \epsilon & 0 & \ldots & 0 & 1-\epsilon\end{array}\right],
\end{equation}

Symmetry flipping:
\begin{equation}
    Q=\left[\begin{array}{ccccc}1-\epsilon & \frac{\epsilon}{n-1} & \cdots & \frac{\epsilon}{n-1} & \frac{\epsilon}{n-1} \\ \frac{\epsilon}{n-1} & 1-\epsilon & \frac{\epsilon}{n-1} & \cdots & \frac{\epsilon}{n-1} \\ \vdots & & \ddots & & \vdots \\ \frac{\epsilon}{n-1} & \cdots & \frac{\epsilon}{n-1} & 1-\epsilon & \frac{\epsilon}{n-1} \\ \frac{\epsilon}{n-1} & \frac{\epsilon}{n-1} & \cdots & \frac{\epsilon}{n-1} & 1-\epsilon\end{array}\right] .
\end{equation}

\section{Software and hardware}
\label{sec:hardware}
We run all experiments with Python 3.8.5 and PyTorch 1.7.0, using NVIDIA TESLA P40 GPUs.

\section{Dataset Details}
\label{sec:app_dataset}
Here we provide the details of graph datasets for node classification.
\begin{table}[!htb]
\centering
% \vspace{-6pt}
\small
% \vspace{-10pt}
% \tabcolsep 0.04in\renewcommand\arraystretch{0.545}{\small{}}
\label{tab:dataset}
% \resizebox*{\linewidth}{!}{%
\begin{tabular}{@{}lrrr@{}}
\toprule
Dataset & $\#$Nodes & $\#$Edges & $\#$Classes \\ \midrule
Cora &  2,485 & 5,069& 7 \\
CiteSeer & 2,110 & 3,668 & 6 \\
PubMed & 19,717& 44,324 & 3 \\
    % DBLP  & 17,716 & 105,734 & 4 \\
    WikiCS & 11,701 & 216,123 & 10 \\
   OGB-arxiv &169,343	&1,166,243& 40\\
\bottomrule
\end{tabular}%}
\label{tab:dataset}
\caption{\small Statistics of the datasets.}
% \vspace{-12pt}
\end{table}

\section{Comparison with traditional graph semi-supervised learning based approaches.}
\label{sec:app_trad}
 For the comparison with the traditional semi-supervised graph embedding methods, we follow the same experimental setting and compare with ICA~\citep{DBLP:conf/icml/LuG03}, Planetoid~\citep{DBLP:conf/icml/YangCS16} and Label Propagation (LP)~\citep{zhuxiaojin} on Cora as follows in Table~\ref{tab:semi_supervised11}. The result shows the advantage of \model across different noise ratios.
\begin{table}[!htb]
  \centering
    \tabcolsep 0.04in\renewcommand\arraystretch{0.9}
% \scriptsize
  \scalebox{0.75}{
    \begin{tabular}{cccccccccc}
   \hline
    Noise type & \multicolumn{5}{c}{Symmetric Noise}   & \multicolumn{4}{c}{Asymmetric Noise} \\
   \hline
    Noise ratio & 0.0   & 0.2   & 0.4   & 0.6   & 0.8   & 0.2   & 0.4   & 0.6   & 0.8  \\
   \hline
    ICA   & 0.729(0.01) & 0.609(0.01) & 0.523(0.04) & 0.394(0.00) & 0.159(0.00) & 0.549(0.00) & 0.453(0.00) & 0.284(0.01) & 0.127(0.01) \\
    LP & 0.603(0.00) & 0.506(0.02) & 0.417(0.03) & 0.297(0.03) & 0.170(0.03) & 0.513(0.03) & 0.391(0.04) & 0.238(0.03) & 0.141(0.02) \\
    Planetoid &   0.739(0.01) & 0.639(0.03) & 0.527(0.04) & 0.379(0.05) & 0.265(0.06) & 0.627(0.03) & 0.441(0.04) & 0.271(0.06) & 0.210(0.09) \\
    \model&\textbf{0.780(0.01)}    &  \textbf{0.739(0.02)}     & \textbf{0.664(0.03) }   &  \textbf{0.515(0.03) }    & \textbf{0.296(0.05) }   & \textbf{ 0.723(0.03)}     & \textbf{0.587(0.07) }   &  \textbf{0.350(0.07) }    &\textbf{ 0.232(0.06)} \\
   \hline
    \end{tabular}}
     \caption{Comparison with more baselines on Cora Dataset.}
     \label{tab:semi_supervised11}%
\end{table}%
% \section{Limitations and social impacts}
% A limitation of our method is that its applicability is limited to exploring one-hop pairwise interactions between nodes, where there exist higher-order pairwise interactions between nodes in the graphs. Moreover, the interactions between the edges are less exploited but may be informative for the GNN to combat the noisy node labels. A potential negative social impact of our work is: the increased computational overhead due to involving multiple GNN models consumes more power energy, which leads to more greenhouse gas emissions.

\section{Experimental results on  heterophilous datasets}
\label{sec:app_het}
We perform extra experiments on heterophilous datasets~\citep{DBLP:journals/corr/abs-2106-06134}. The results are demonstrated in the Table~\ref{tab:hetro_results}. It shows that \model is still able to outperform the vanilla one except for one case in Chameleon dataset. 
% Added by Roy
%This improvement may due to the decoupling operation that the PI label is learned from both the graph connectivity and node feature. 
% 
%
Meanwhile, the improvement is somewhat smaller, which implies \model may be more effective on homophilous datasets. 
\begin{table}[!htb]
    \centering
    \small
      \tabcolsep 0.04in\renewcommand\arraystretch{0.9}
    \begin{tabular}{c|cc|cc}
    \hline
       Noise type  & \multicolumn{2}{c}{Symmetric} & \multicolumn{2}{c}{Asymmetric} \\
       \hline
     &  \multicolumn{4}{c}{Actor}\\
       \hline
       Noise Ratio  &0.4&0.6& 0.4&0.6  \\
        \hline
        GCN&0.209(0.02) & 0.208(0.02)&0.198(0.03) & 0.199(0.02) \\
        \model w/o pc & 0.216(0.01) & 0.210(0.02)&0.201(0.02)  & \textbf{0.202(0.02)}\\
        \model& \textbf{0.218(0.02)} & \textbf{0.213(0.02)} &\textbf{0.204(0.02)}& 0.200(0.02)\\
       \hline
     & \multicolumn{4}{c}{Chameleon}\\
     \hline
      GCN&0.251(0.03) & 0.246(0.03) &  \textbf{0.245(0.04)} & 0.228(0.03) \\
         \model w/o pc &0.264(0.03)  & 0.249(0.03)& 0.242(0.05) & 0.229(0.04) \\
         \model& \textbf{0.269(0.02)} & \textbf{0.251(0.03)} &0.239(0.05) &\textbf{0.237(0.04)} \\
         \hline
    \end{tabular}
    \caption{Experimental results on  heterophilous datasets.}
    \label{tab:hetro_results}
\end{table}

\section{Significance test results}
\label{sec:app_sig}
We  perform  significance  tests  to  verify  whether  \model  outperforms  the  vanilla GNN model significantly  using double-sided T-test in Table~\ref{tab:tests_cifar10}.  We use python package ``scipy.stats.ttest1samp" and report the average results over 10 different runs as follows. \model is  better than  GAT because  the  absolute  value  of  the  t-statistic  is relatively large and the p-value is small.
\begin{table}[!htb]
\small
    \centering
     
    \begin{tabular}{c|ccc}
    \hline
         Method& Setting& $|$T-statistic$|$&p-value  \\
         \hline
        & \multicolumn{3}{c}{Cora}\\
         \hline
       \multirow{2}*{GAT vs. \model}& Symmetric Noise-0.8&4.78&0.001\\
         & Asymmetric Noise-0.8& 3.42& 0.008\\
         \hline
            & \multicolumn{3}{c}{CiteSeer}\\
            \hline
  \multirow{2}*{GAT vs. \model}    & Symmetric Noise-0.8 &  2.09&0.060\\
       & Asymmetric Noise-0.8  & 4.63&0.001 \\
         \hline
    \end{tabular}
   \caption{Statistical significance tests.}
    \label{tab:tests_cifar10}
\end{table}

\section{Experimental results on using clean label comparison}
\label{sec:app_clean}
To observe the node classification results by training with PI labels from clean label comparison (which are obtained by comparing the clean class labels for two nodes), we did experiments on Cora, CiteSeer and PubMed with GCN and a noise ratio of 0.4 and 0.6. The results are shown in the Table~\ref{tab:clean_label}. In most cases, clean label comparison can help the \model to combat noisy labels except for some challenging cases with asymmetric noise. One reason may be the inherent noise exists in clean node labels for Cora, where we cannot obtain perfectly clean PI label.

% perfect pI labels are not obtainable. 
\begin{table}[!htb]
\small
    \centering
 \tabcolsep 0.04in\renewcommand\arraystretch{0.9}
    \begin{tabular}{c|cc|cc}
    \hline
       Noise type  & \multicolumn{2}{c}{Symmetric} & \multicolumn{2}{c}{Asymmetric} \\
       \hline
     &  \multicolumn{4}{c}{Cora}\\
       \hline
       Noise Ratio  &0.4&0.6& 0.4&0.6  \\
        \hline
         \model& 0.664(0.03)& 0.515(0.03)&0.587(0.07) &\textbf{0.350(0.07)} \\
       Clean  \model  & \textbf{0.671(0.03)}&\textbf{0.523(0.03)}&\textbf{0.589(0.07)}&0.341(0.07)\\
       \hline
     & \multicolumn{4}{c}{CiteSeer}\\
     \hline
       \model&0.591(0.03) &0.432(0.07)& 0.531(0.06) &0.353(0.06)  \\
         Clean  \model  &\textbf{0.605(0.04)} & \textbf{0.447(0.07)}& \textbf{0.536(0.05)} &\textbf{0.355(0.05)} \\
         \hline
            & \multicolumn{4}{c}{PubMed}\\
            \hline
          \model&   0.638(0.04) &0.470(0.08) 
            &0.583(0.07) &0.425(0.07)\\
             Clean  \model  & \textbf{0.640(0.02)} &\textbf{0.485(0.07) }
            &\textbf{0.590(0.07)} &\textbf{0.429(0.07)}\\
     \hline
    \end{tabular}
        \caption{Experimental results on using clean label comparison. Clean \model means the \model is trained with the PI labels from clean label comparison.}
    \label{tab:clean_label}
\end{table}

\section{Experimental results with lower noise ratios}
\label{sec:app_small_ratio}
For the \model under lower noise ratios, we empirically verify its effectiveness on Cora and CiteSeer with the noise ratio of 0.1, which is shown in Table~\ref{tab:small_ratio}.

\begin{table}[!htb]
\small
    \centering
    
    \begin{tabular}{c|c|c}
    \hline
       Noise type  & \multicolumn{1}{c}{Symmetric} & \multicolumn{1}{c}{Asymmetric} \\
       \hline
     &  \multicolumn{2}{c}{Cora}\\
       \hline
       Noise Ratio  &0.1& 0.1  \\
        \hline
         GCN& 0.766(0.03)&0.762(0.04) \\
        \model w/o pc & 0.769(0.03)&0.763(0.03)\\
       \model &\textbf{0.772(0.02)} &\textbf{0.768(0.03)}\\
     \hline
       &  \multicolumn{2}{c}{CiteSeer}\\
       \hline
       Noise Ratio  &0.1& 0.1  \\
        \hline
         GCN&0.642(0.03)  &0.618(0.05)\\
        \model w/o pc & 0.648(0.04) & 0.633(0.02)\\
       \model & \textbf{0.659(0.03)}&\textbf{0.658(0.05)}\\
       \hline
    \end{tabular}
    \caption{Experimental results with lower noise ratios.}
    \label{tab:small_ratio}
\end{table}

\section{Comparison of the training time}
\label{sec:training_time}
We compare the training time of \model and the vanilla GNN as follows. We observe that using dual GNNs does not incur much higher computational cost because the two GNNs run in parallel rather than in sequence on the GPU and the major time-consuming part is for data loading and transforms rather than forward/backward pass. Additionally, \model does not incur more inference cost than a vanilla model.
% \vspace{-1em}
\begin{table}[h]
    \centering
    % \scriptsize
    \begin{tabular}{c|ccccc}
    \hline
       Dataset  & Cora & CiteSeer& PubMed& WikiCS& OGB \\
       \hline
       Time-GCN(s)  &25.79  & 26.98 & 140.53 & 71.12 & 3930.14\\
      Time-\model(s)   &  26.46& 30.33 & 160.29 &  81.95& 4117.00\\
      
      \hline
    \end{tabular}
\end{table}

\section{Additional comparison with baselines}
\label{sec:baselines_appendix}
In addition to Table~\ref{tab:baselines} in the main paper, we provide the comparison on PubMed and WikiCS with baselines to further demonstrate the effectiveness of our \model, which is shown as follows. 
% \vspace{-0.8em}
\begin{table}[h]
% \tiny
% \scriptsize
  \centering
%   \tabcolsep 0.04in\renewcommand\arraystretch{0.9}{\small{}}%
%   \fontsize{7.8pt}{7.8pt}\selectfont
    \begin{tabular}{c|c|c}
    \hline
    Noise type & Symmetric Noise(0.4)   & Asymmetric Noise(0.4) \\
   \hline
   & \multicolumn{2}{c}{Test dataset: PubMed / WikiCS } \\
 \hline 
  Decoupling&0.627(0.05) / 0.555(0.05) & 0.578(0.06) / 0.465(0.05) \\
     GCE& 0.604(0.06) / 0.536(0.06) & 0.524(0.08) / 0.456(0.06)\\
    APL &  0.606(0.06) / 0.526(0.09) & 0.524(0.08) / 0.336(0.11) \\
   Co-teaching& 0.523(0.06) / 0.337(0.10)& 0.433(0.11) / 0.262(0.08) \\
     LPM-1& 0.634(0.06) / 0.554(0.04) & 0.570(0.09) / 0.401(0.04)\\
   T-Revision&0.603(0.04) / 0.542(0.07)& 0.554(0.08) / 0.478(0.10) \\
   DivideMix& 0.543(0.08) / 0.419(0.08) & 0.566(0.07) / 0.169(0.08) \\
   \hline
    \model (ours)&\textbf{0.638(0.04)} / \textbf{0.562(0.04) }& \textbf{0.583(0.07)} / \textbf{0.483(0.05)}\\
    \hline 
 \end{tabular}
  \label{tab:baselines_appendix}%
  \caption{Additional comparison with baselines on two different datasets. We use the GCN as the graph neural network backbone.}
%   \vspace{-1.8em}
\end{table}%

\newpage
\section{\textcolor{black}{Sensitivity to the model initialization}}
\textcolor{black}{We provide the sensitivity analysis for \model on different model initializations in Table~\ref{tab:sensi_init}.  The results demonstrate that the performance is not sensitive to different model initializations, where the biggest gap among all the model initializations is smaller than 3\%.}
\begin{table*}[!h]
% \tiny
% \scriptsize
  \centering

  % \tabcolsep 0.04in\renewcommand\arraystretch{0.9}{\small{}}%

  % \fontsize{7.8pt}{7.8pt}\selectfont
   \scalebox{1.0}{ 
      \color{black}\begin{tabular}{c|cc}
    % \hline
    Noise type &Symmetric Noise   & Asymmetric Noise \\
    \hline
 
    Noise ratio &  0.6   & 0.6      \\
   \hline
   & \multicolumn{2}{c}{Test dataset: Cora / CiteSeer} \\

 \hline
 
 Uniform Initialization& 0.514(0.05) / 0.443(0.05) & 0.364(0.02) / 0.321(0.04) \\
  Normal Initialization& 0.548(0.03) / 0.426(0.05) & 0.361(0.05) / 0.345(0.01) \\
  Constant Initialization & 0.504(0.01) / 0.413(0.04) & 0.339(0.03) / 0.341(0.04) \\
   Kaiming Initialization& 0.503(0.04) / 0.429(0.03) & 0.333(0.02) / 0.372(0.09) \\
Glorot Initialization (Ours) & 0.515(0.03) / 0.432(0.07) & 0.350(0.07) / 0.353(0.06) \\
   \hline
  % \rowcolor{Gray}    \model (ours)& \textbf{0.664(0.03) / 0.591(0.03)} & \textbf{0.515(0.03) / 0.432(0.07)} &  \textbf{0.587(0.07) / 0.531(0.06)} & \textbf{0.350(0.07)} / \textbf{0.353(0.06)}\\
    
    % \hline 
 \end{tabular}}
 \caption{\small  \textcolor{black}{Sensitivity to the model initialization. Model architecture is the GCN.}}
  \label{tab:sensi_init}%
  % \vspace{-1.5em}
\end{table*}%

\section{\textcolor{black}{Results on a larger graph dataset}}
\textcolor{black}{We evaluate our proposed \model on an even larger OGB-products dataset, which has 2,449,029 nodes with 61,859,140 edges and thus is much larger than the OGB-arxiv dataset (169,343 nodes and 1,166,243 edges) used in Table~\ref{tab:effectiveness}. The results are shown in Table~\ref{tab:ogb_product}, where our \model can still demonstrate promise compared to the vanilla GCN model and \model w/o predictive confidence.}

\begin{table*}[!h]
    \centering
    \tabcolsep 0.04in\renewcommand\arraystretch{0.745}{\small{}}%
    \scriptsize
    % \vspace{-0.5em}
     
    \resizebox{0.99\textwidth}{!}{%
    \color{black}\begin{tabular}{cccccccccc}
\hline
\multicolumn{1}{c|}{Noise type}    & \multicolumn{1}{c|}{\cellcolor[HTML]{ECF4FF}No Noise}             & \multicolumn{4}{c|}{Symmetric Noise}                                                                           & \multicolumn{4}{c}{Asymmetric Noise}                                                     \\ \hline
\multicolumn{10}{c}{OGB-products}                                                                                                                                                                                                                                                                                           \\ \hline
\multicolumn{1}{c|}{Noise ratio}   & \multicolumn{1}{c|}{\cellcolor[HTML]{ECF4FF}0.0}                  & 0.2                  & 0.4                  & 0.6                  & \multicolumn{1}{c|}{0.8}                  & 0.2                  & 0.4                  & 0.6                  & 0.8                  \\ \hline
\multicolumn{1}{c|}{GCN}           & \multicolumn{1}{c|}{\cellcolor[HTML]{ECF4FF}\textbf{0.732(0.04)}} & 0.709(0.01)          & 0.661(0.09)          & 0.612(0.07)          & {0.471(0.06)}          & 0.689(0.03)          & 0.635(0.03)          & 0.102(0.03)          & 0.053(0.01)          \\
\multicolumn{1}{c|}{\model w/o pc} & \multicolumn{1}{c|}{\cellcolor[HTML]{ECF4FF}0.711(0.03)}          & 0.721(0.04) & 0.669(0.03)          & 0.631(0.06)          & {0.487(0.09)}          & 0.700(0.09)          & 0.641(0.04)          & 0.136(0.01) & 0.110(0.04) \\
\multicolumn{1}{c|}{\model}        & \multicolumn{1}{c|}{\cellcolor[HTML]{ECF4FF}0.727(0.05)}     & \textbf{0.738(0.06)}    & \textbf{0.677(0.06)}          & \textbf{0.658(0.03)} & \textbf{0.506(0.05)}  & \textbf{0.719(0.06)} & \textbf{0.667(0.03)} & \textbf{0.196(0.06)}          & \textbf{0.153(0.04)}          \\ \hline

\end{tabular}
}
\caption{\small \textcolor{black}{Test accuracy on the OGB-products dataset for \model with GCN as the backbone. \textbf{Bold} numbers are superior results. Std. is shown in the bracket. w/o pc means that the PI labels are not estimated using the predictive confidence but just the node connectivity.}}
% \vspace{-4ex}
    \label{tab:ogb_product}
\end{table*}

\section{\textcolor{black}{Results on using low-rank approximation for PI estimation}}
\textcolor{black}{We estimate the PI labels by performing SVD on the input graph and using the low-rank representations as the estimation results. Suppose the rank of the representation is $r$, we tested the node classification performance under different values of $r$ (i.e., 1, 50, 100, 200, 400, 600, 1000).  Note that we use GCN and Cora as the GNN architecture and the dataset, respectively. During the low-rank approximation, we force the values of the smoothed reconstruction to be greater than 0 and less than 1 by value clipping. The results are updated in Table~\ref{tab:low_rank}.}

\begin{table*}[!h]
    \centering
    \tabcolsep 0.04in\renewcommand\arraystretch{0.745}{\small{}}%
    \scriptsize
    % \vspace{-0.5em}
     
    \resizebox{0.99\textwidth}{!}{%
    \color{black}\begin{tabular}{c|cccc|cccc}
\hline
\multicolumn{1}{c|}{Noise type}                & \multicolumn{4}{c|}{Symmetric Noise}                                                                           & \multicolumn{4}{c}{Asymmetric Noise}                                                     \\ \hline        
{Noise ratio}                    & 0.2                  & 0.4                  & 0.6                  & \multicolumn{1}{c|}{0.8}                  & 0.2                  & 0.4                  & 0.6                  & 0.8                  \\ \hline
{GCN}            & 0.722(0.03)          & 0.613(0.07)          & 0.446(0.06)          & {0.285(0.07)}          & 0.703(0.04)          & 0.514(0.06)          & 0.291(0.04)          & 0.161(0.02)       \\
{Low-rank approximation-1}           & 0.693(0.01)&0.587(0.03)&0.418(0.05)&0.272(0.03)&0.688(0.06)&0.483(0.04)&0.269(0.02) &0.110(0.05)\\
Low-rank approximation-50 &0.698(0.04) &0.593(0.03)&0.436(0.03)&0.269(0.01)&0.679(0.04)&0.502(0.06)&0.271(0.08) &0.132(0.08) \\
Low-rank approximation-100 & 0.712(0.09) &0.603(0.05)&0.427(0.06)&0.277(0.05)&0.673(0.01)&0.515(0.06)&0.286(0.04) &0.149(0.02) \\
Low-rank approximation-200 &  0.732(0.04) &0.607(0.04)&0.474(0.04)&0.282(0.03)&0.694(0.05)&0.547(0.06)&0.294(0.01) &0.173(0.08) \\
Low-rank approximation-400 & 0.734(0.01) &0.611(0.09)&0.496(0.05)&\textbf{0.300(0.07)}&0.686(0.01)&0.563(0.02)&0.319(0.00) &0.183(0.05) \\
Low-rank approximation-600 & 0.726(0.02) &0.626(0.02)&0.476(0.03)&0.298(0.03)&0.695(0.04)&0.534(0.07)&0.321(0.03) &0.201(0.03) \\
Low-rank approximation-1000 &0.704(0.05) &0.610(0.08)&0.455(0.04)&0.256(0.07)&0.668(0.02)&0.508(0.05)&0.346(0.07) &0.188(0.09) \\
{\model}             & \textbf{0.739(0.02)}          & \textbf{0.664(0.03)} & \textbf{0.515(0.03)} & 0.296(0.05) & \textbf{0.723(0.03)} & \textbf{0.587(0.07)} & \textbf{0.350(0.07)}          & \textbf{0.232(0.06)}           \\ \hline

\end{tabular}
}
\caption{\small \textcolor{black}{Test accuracy on using low-rank approximation as the PI labels.}}
% \vspace{-4ex}
    \label{tab:low_rank}
\end{table*}

\textcolor{black}{$r=400$ roughly achieves the best performance but still cannot outperform our \model. The reason might be that low-rank approximation is mainly designed for matrix compression, denoising and completion (https://web.stanford.edu/class/cs168/l/l9.pdf), which cannot capture the uncertainty of the PI labels in essence as in \model. Moreover, we observe that during low-rank approximation, the values of each node pair will quickly goes from negative values to values larger than 1, which is not suitable to be the PI labels for training. Although we directly adopt the clipping approach to force the values to be in the range of 0 and 1, additional curated designs might be more beneficial. Finally, the rank $r$ is an important hyperparameter to tune in the low-rank approximation approaches, which requires extra manual tuning compared to our \model. }

\section{\textcolor{black}{Discusssion on a different training strategy}}
\textcolor{black}{In Table~\ref{tab:alter_train_cora} and~\ref{tab:alter_train_cite}, we test the performance of \model using a different training scheme, where we firstly pretrain the PI label estimation network $f_e$ for 400 epochs. Then we directly apply the pretrained $f_e$ to output the estimated PI labels for training $f_t$. We show the results on Cora and CiteSeer datasets as follows:}

\begin{table*}[!h]
    \centering
    \tabcolsep 0.04in\renewcommand\arraystretch{0.745}{\small{}}%
    \scriptsize
    % \vspace{-0.5em}
     
    \resizebox{0.99\textwidth}{!}{%
    \color{black}\begin{tabular}{cccccccccc}
\hline
\multicolumn{1}{c|}{Noise type}    & \multicolumn{1}{c|}{\cellcolor[HTML]{ECF4FF}No Noise}             & \multicolumn{4}{c|}{Symmetric Noise}                                                                           & \multicolumn{4}{c}{Asymmetric Noise}                                                     \\ \hline
\multicolumn{10}{c}{OGB-products}                                                                                                                                                                                                                                                                                           \\ \hline
\multicolumn{1}{c|}{Noise ratio}   & \multicolumn{1}{c|}{\cellcolor[HTML]{ECF4FF}0.0}                  & 0.2                  & 0.4                  & 0.6                  & \multicolumn{1}{c|}{0.8}                  & 0.2                  & 0.4                  & 0.6                  & 0.8                  \\ \hline
\multicolumn{1}{c|}{\model (Pretrain)}           & \multicolumn{1}{c|}{\cellcolor[HTML]{ECF4FF}0.772(0.01)} & 0.733(0.03) & \textbf{0.672(0.04)} &  0.508(0.06) & \textbf{0.319(0.04) }& \textbf{0.728(0.04)} & 0.569(0.04) & 0.339(0.06) & \textbf{0.245(0.06) }        \\
\multicolumn{1}{c|}{\model (Ours)} & \multicolumn{1}{c|}{\cellcolor[HTML]{ECF4FF}\textbf{0.780(0.01)}}  & \textbf{0.739(0.02)}& 0.664(0.03)& \textbf{0.515(0.03) }& 0.296(0.05) &0.723(0.03) &\textbf{0.587(0.07) }& \textbf{0.350(0.07)} & 0.232(0.06)\\

\hline
\end{tabular}
}

\caption{\small \textcolor{black}{Test accuracy on Cora using a different training scheme.}}
% \vspace{-4ex}
    \label{tab:alter_train_cora}
\end{table*}

\begin{table*}[!h]
    \centering
    \tabcolsep 0.04in\renewcommand\arraystretch{0.745}{\small{}}%
    \scriptsize
    % \vspace{-0.5em}
     
    \resizebox{0.99\textwidth}{!}{%
    \color{black}\begin{tabular}{cccccccccc}
\hline
\multicolumn{1}{c|}{Noise type}    & \multicolumn{1}{c|}{\cellcolor[HTML]{ECF4FF}No Noise}             & \multicolumn{4}{c|}{Symmetric Noise}                                                                           & \multicolumn{4}{c}{Asymmetric Noise}                                                     \\ \hline
\multicolumn{10}{c}{OGB-products}                                                                                                                                                                                                                                                                                           \\ \hline
\multicolumn{1}{c|}{Noise ratio}   & \multicolumn{1}{c|}{\cellcolor[HTML]{ECF4FF}0.0}                  & 0.2                  & 0.4                  & 0.6                  & \multicolumn{1}{c|}{0.8}                  & 0.2                  & 0.4                  & 0.6                  & 0.8                  \\ \hline
\multicolumn{1}{c|}{\model (Pretrain)}           & \multicolumn{1}{c|}{\cellcolor[HTML]{ECF4FF}\textbf{0.693(0.01)}} &    0.631(0.03) &\textbf{0.606(0.04)} & \textbf{0.452(0.04)}& 0.229(0.05)&0.604(0.04) & 0.511(0.04)& 0.338(0.02) & \textbf{0.242(0.06)}   \\
\multicolumn{1}{c|}{\model (Ours)} & \multicolumn{1}{c|}{\cellcolor[HTML]{ECF4FF}0.684(0.03)} &\textbf{0.642(0.03)} &0.591(0.03)  &0.432(0.07) & \textbf{0.245(0.05)} &\textbf{0.628(0.03)}&\textbf{0.531(0.06)}&\textbf{0.353(0.06)}&0.238(0.06) \\

\hline
\end{tabular}
}
\caption{\small \textcolor{black}{Test accuracy on CiteSeer using a different training scheme.}}
% \vspace{-4ex}
    \label{tab:alter_train_cite}
\end{table*}

\textcolor{black}{where we can observe that firstly pretraining the PI label estimation network and freezing it during training the node classification model achieves a similar classification performance compared to our training scheme (\emph{c.f.} Algorithm~\ref{alg:algo}).}

\section{ \textcolor{black}{Additional cases of the teaser example}}
\label{sec:app_corner_case}
\textcolor{black}{We note that there are additional cases that the conclusion of the teaser example might not hold. For instance, four nodes have the clean labels as $x1-0,x2-0,x3-0,x4-0$, after noise corruption, the node labels are changed to $x1-1,x2-0,x3-0, x4-0$. Therefore, the noise ratio of the node labels is 25\% while that of the PI labels is 37.5\%. We provide the comparison of noise ratios between PI labels and node labels on real-world datasets in Section~\ref{sec:app_noise_ratio_comparison} in order to verify the validity of our proposed \model.}

\section{ \textcolor{black}{Noise ratio comparison between PI labels and node labels on  real-world datasets}}
\label{sec:app_noise_ratio_comparison}
\textcolor{black}{For evaluating the noise ratio of the PI labels on real-world graph datasets, we compare the noise ratio of the PI labels and node labels on Cora and CiteSeer datasets. The results are shown as follows: }
\begin{table}[!h]
    \centering
    \color{black}\begin{tabular}{c|cccc}
    \hline
    
\multicolumn{5}{c}{Cora} \\
    \hline
       Noise ratio of the node labels (Symmetric)   & 0.2 & 0.4 & 0.6 & 0.8 \\
       Noise ratio of the PI labels &0.008&	0.021&	0.031&	0.036 \\
       \hline
        Noise ratio of the node labels (Asymmetric)   & 0.2 & 0.4 & 0.6 & 0.8 \\
       Noise ratio of the PI labels & 0.012&	0.021&	0.029	&0.035\\
       \hline
       \multicolumn{5}{c}{CiteSeer} \\
       \hline
         Noise ratio of the node labels (Symmetric)   & 0.2 & 0.4 & 0.6 & 0.8 \\
       Noise ratio of the PI labels & 0.009&	0.017	&0.024	&0.034\\
       \hline
        Noise ratio of the node labels (Asymmetric)   & 0.2 & 0.4 & 0.6 & 0.8 \\
       Noise ratio of the PI labels & 0.009	&0.016	&0.025&	0.031\\
         \hline
    \end{tabular}
    \caption{\small \textcolor{black}{Noise ratio comparison between PI labels and node labels on  real-world datasets}}
    \label{tab:app_temp}
\end{table}

\textcolor{black}{From the above table, we can observe the noise ratio of the PI labels is indeed small on the real graph datasets, which is able to justify the intuition of our \model.}

\section{ \textcolor{black}{Subgraph sampling on smaller graphs}}
\textcolor{black}{We test the performance of \model (GCN as the backbone) after using subgraph sampling on smaller graphs, i.e., Cora and CiteSeer datasets. Specifically, we sample 15 and 10 neighbors for each node in the 1st and 2nd layer of the GNN and
set the batch size to 128. The results are shown as follows:}

\begin{table*}[!h]
    \centering
    \tabcolsep 0.04in\renewcommand\arraystretch{0.745}{\small{}}%
    \scriptsize
    % \vspace{-0.5em}
     
    \resizebox{0.99\textwidth}{!}{%
    \color{black}\begin{tabular}{cccccccccc}
\hline
\multicolumn{1}{c|}{Noise type}    & \multicolumn{1}{c|}{\cellcolor[HTML]{ECF4FF}No Noise}             & \multicolumn{4}{c|}{Symmetric Noise}                                                                           & \multicolumn{4}{c}{Asymmetric Noise}                                                     \\ \hline
\multicolumn{10}{c}{OGB-products}                                                                                                                                                                                                                                                                                           \\ \hline
\multicolumn{1}{c|}{Noise ratio}   & \multicolumn{1}{c|}{\cellcolor[HTML]{ECF4FF}0.0}                  & 0.2                  & 0.4                  & 0.6                  & \multicolumn{1}{c|}{0.8}                  & 0.2                  & 0.4                  & 0.6                  & 0.8                  \\ \hline
\multicolumn{1}{c|}{\model (Subgraph sampling)}           & \multicolumn{1}{c|}{\cellcolor[HTML]{ECF4FF}0.771(0.01)} & 0.728(0.03) & 0.651(0.03) & \textbf{0.523(0.05)} & \textbf{0.311(0.02)}&0.708(0.03) &  0.573(0.07) & 0.349(0.01) & \textbf{0.246(0.07)}      \\
\multicolumn{1}{c|}{\model (Ours)} & \multicolumn{1}{c|}{\cellcolor[HTML]{ECF4FF}\textbf{0.780(0.01)}}  & \textbf{0.739(0.02)}& \textbf{0.664(0.03)}& 0.515(0.03) & 0.296(0.05) &\textbf{0.723(0.03)} &\textbf{0.587(0.07) }& \textbf{0.350(0.07)} & 0.232(0.06)\\

\hline
\end{tabular}
}

\caption{\small \textcolor{black}{Test accuracy on Cora using subgraph sampling.}}
% \vspace{-4ex}
    \label{tab:sub_cora}
\end{table*}

\begin{table*}[!h]
    \centering
    \tabcolsep 0.04in\renewcommand\arraystretch{0.745}{\small{}}%
    \scriptsize
    % \vspace{-0.5em}
     
    \resizebox{0.99\textwidth}{!}{%
    \color{black}\begin{tabular}{cccccccccc}
\hline
\multicolumn{1}{c|}{Noise type}    & \multicolumn{1}{c|}{\cellcolor[HTML]{ECF4FF}No Noise}             & \multicolumn{4}{c|}{Symmetric Noise}                                                                           & \multicolumn{4}{c}{Asymmetric Noise}                                                     \\ \hline
\multicolumn{10}{c}{OGB-products}                                                                                                                                                                                                                                                                                           \\ \hline
\multicolumn{1}{c|}{Noise ratio}   & \multicolumn{1}{c|}{\cellcolor[HTML]{ECF4FF}0.0}                  & 0.2                  & 0.4                  & 0.6                  & \multicolumn{1}{c|}{0.8}                  & 0.2                  & 0.4                  & 0.6                  & 0.8                  \\ \hline
\multicolumn{1}{c|}{\model (Subgraph sampling)}           & \multicolumn{1}{c|}{\cellcolor[HTML]{ECF4FF}0.674(0.01)} &0.639(0.04) &  0.579(0.07)  &0.430(0.09) &\textbf{0.258(0.05)} & 0.618(0.04) & 0.528(0.04) & 0.348(0.03)  & \textbf{0.250(0.08)} \\
\multicolumn{1}{c|}{\model (Ours)} & \multicolumn{1}{c|}{\cellcolor[HTML]{ECF4FF}\textbf{0.684(0.03)}} &\textbf{0.642(0.03)} &\textbf{0.591(0.03)}  &\textbf{0.432(0.07)} & 0.245(0.05) &\textbf{0.628(0.03)}&\textbf{0.531(0.06)}&\textbf{0.353(0.06)}&0.238(0.06) \\

\hline
\end{tabular}
}
\caption{\small \textcolor{black}{Test accuracy on CiteSeer using  subgraph sampling.}}
% \vspace{-4ex}
    \label{tab:sub_cite}
\end{table*}

\textcolor{black}{where we can observe that applying subgraph sampling is less effective than using the entire graph as the input on smaller noise ratios. }

\section{ \textcolor{black}{Comparison with baselines on OGB-arxiv}}
\textcolor{black}{In Table~\ref{tab:app_compara_ogb}, we implement all the baselines on OGB-arxiv dataset and compare them with \model using subgraph sampling.}

\begin{table*}[!h]
% \tiny
% \scriptsize
  \centering

  % \tabcolsep 0.04in\renewcommand\arraystretch{0.9}{\small{}}%
  
  % \fontsize{7.8pt}{7.8pt}\selectfont
   \scalebox{1.0}{ 
    \color{black}\begin{tabular}{c|cc|cc}
    % \hline
    Noise type & \multicolumn{2}{c}{Symmetric Noise}   & \multicolumn{2}{c}{Asymmetric Noise} \\
    \hline
    Noise ratio &  0.4   & 0.6     & 0.4 & 0.6  \\
   \hline
   & \multicolumn{4}{c}{Test dataset: OGB-arxiv} \\

 \hline
  Decoupling& 0.385(0.09) & 0.347(0.05) & 0.411(0.04) & 0.029(0.01)\\
     GCE&0.451(0.05) & 0.407(0.02) & 0.391(0.06) & 0.057(0.03) \\
    APL & 0.412(0.05) & 0.375(0.05) & 0.399(0.06) & 0.062(0.01) \\
   Co-teaching&0.461(0.04) & 0.403(0.04) & 0.410(0.05) & 0.038(0.01) \\
     LPM-1&0.450(0.01) & 0.397(0.03) & 0.439(0.06) & 0.056(0.01)\\
   T-Revision&  0.417(0.04) &  0.409(0.05) & 0.427(0.05) & \textbf{0.071(0.04)} \\
   DivideMix& 0.448(0.01) & 0.403(0.05) & 0.438(0.05) & 0.041(0.02)\\
   \hline
  \rowcolor{Gray}    \model (ours)& \textbf{0.467(0.03)} &\textbf{0.418(0.04)}  &\textbf{0.461(0.01)} &0.069(0.01)\\
    
    % \hline 
 \end{tabular}}
 \caption{\small \textcolor{black}{Comparative results with baselines on OGB-arxiv.}}
  \label{tab:app_compara_ogb}%
  % \vspace{-1.5em}
\end{table*}%
\textcolor{black}{where we observe \model can still outperform all the baselines in most cases.}

\end{document}